%% file: rnlvr.tex
\documentclass[11pt,a4paper]{article}
\usepackage[hyperref]{acl2019}
\usepackage{times,latexsym}
\usepackage{url}
\usepackage[T1]{fontenc}
\usepackage{xr}

\aclfinalcopy %
\def\aclpaperid{809} %

\usepackage{xspace,mfirstuc,tabulary}

\usepackage{amsmath}
\usepackage{multirow}
\usepackage{adjustbox}
\usepackage{balance}
\usepackage{url}
\usepackage{color}
\usepackage{array}
\usepackage{arydshln}
\usepackage{tikz,pgfplots}
\usepackage{pifont}
\usepackage{xspace}
\usepackage{setspace}

\DeclareSymbolFont{extraup}{U}{zavm}{m}{n}
\DeclareMathSymbol{\varheart}{\mathalpha}{extraup}{86}
\DeclareMathSymbol{\vardiamond}{\mathalpha}{extraup}{87}
\newcommand{\affilone}{$^\ddagger$}

\newcommand{\eat}[1]{}
\newcommand{\stdev}[2]{${#1}$\begin{footnotesize}$\pm{#2}$\end{footnotesize}}

\usepackage{titlesec}
\titlespacing{\paragraph} {0pt}{0.3ex plus 1ex minus .2ex}{1em}

\newcommand{\dline}{\hdashline[0.5pt/1pt]}

\newcommand{\realnlvr}{NLVR2\xspace}

\newcommand{\nlstring}[1]{{\em #1}}
\newcommand{\synset}[1]{{\tt #1}\xspace}
\newcommand{\system}[1]{\textsc{#1}\xspace}
\newcommand{\film}{FiLM\xspace}

\newcommand{\truelabel}{\mathrm{True}}
\newcommand{\falselabel}{\mathrm{False}}

\author{Alane Suhr\affilone\thanks{\hspace{5pt}Contributed equally.},\hspace{1pt} Stephanie Zhou\thanks{\hspace{5pt}Work done as an undergraduate at Cornell University.}\hspace{4pt},\hspace{-3pt}\footnotemark[1] \hspace{3.5pt}Ally Zhang\affilone, Iris Zhang\affilone, Huajun Bai\affilone, {\normalfont and} Yoav Artzi\affilone
\vspace{0.5em}\\
 \affilone Cornell University  Department of Computer Science and Cornell Tech \\ %
  New York, NY 10044 \\
  {\tt \{suhr, yoav\}@cs.cornell.edu \hspace{1em}\{az346, wz337, hb364\}@cornell.edu}
  \vspace{0.5em}\\
   $^\dagger$University of Maryland Department of Computer Science\\
 College Park, MD 20742 \\
  {\tt stezhou@cs.umd.edu}}

\date{}

\title{A Corpus for Reasoning About Natural Language \\ Grounded in Photographs}
\begin{document}

\maketitle
\begin{abstract}	
\input{abstract}
\end{abstract}

\input{introduction}

\input{related}

\input{collection}

\input{analysis}
\input{human_performance}
\input{evaluation_systems}
\input{results}

\input{discussion}

\input{ack}

\balance
\bibliography{main}
\bibliographystyle{acl_natbib}

\clearpage
\nobalance
\appendix
\input{supplementary}

\end{document}

%% file: abstract.tex
We introduce a new dataset for joint reasoning about natural language and images, with a focus on semantic diversity, compositionality, and visual reasoning challenges. The data contains $107{,}292$ examples of English sentences paired with web photographs. The task is to determine whether a natural language caption is true about a pair of photographs. We crowdsource the data using sets of visually rich images and a compare-and-contrast task to elicit linguistically diverse language. Qualitative analysis shows the data requires compositional joint reasoning, including about quantities, comparisons, and relations. Evaluation using state-of-the-art visual reasoning methods shows the data presents a strong challenge.

%% file: introduction.tex
\section{Introduction}
\label{sec:intro}

Visual reasoning with natural language is a promising avenue to study compositional semantics by grounding words, phrases, and complete sentences to objects, their properties, and relations in images. 
This type of linguistic reasoning is critical for  interactions grounded in visually complex environments, such as in robotic applications. %
However, commonly used resources for language and vision~\cite[e.g.,][]{Antol:15vqa,Chen:16coco} focus mostly on identification of object properties and few spatial relations~\cite[Section~\ref{sec:analysis};][]{Ferraro:15,Alikhani:19}.
This relatively simple reasoning, together with biases in the data, removes much of the need to consider language compositionality~\cite{Goyal:17}. 
This motivated the design of datasets that require compositional\footnote{In parts of this paper, we use the term \emph{compositional} differently than it is commonly used in linguistics to refer to reasoning that requires composition. This type of reasoning often manifests itself in highly compositional language.} visual reasoning, including NLVR~\cite{Suhr:17visual-reason} and CLEVR~\cite{Johnson:16clevr,Johnson:17iep}. 
These datasets use synthetic images, synthetic language, or both. 
The result is a limited representation of linguistic challenges: 
synthetic languages are inherently of bounded expressivity, 
and synthetic visual input entails limited lexical and semantic diversity.

\begin{figure}[t]
	\centering\footnotesize
	\fbox{
		\begin{minipage}{0.95\linewidth}
			\centering
			\frame{\colorbox{gray!20}{\frame{\includegraphics[height=16ex]{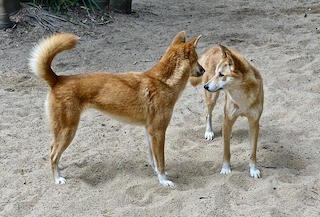}}~{\frame{\includegraphics[height=16ex]{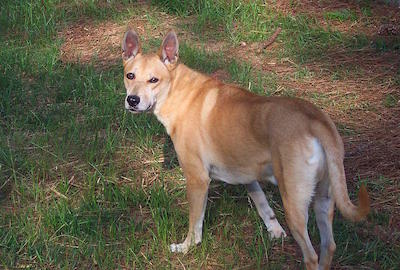}}}}} \\
			\begin{footnotesize}
			\nlstring{The left image contains twice the number of dogs as the right image, and at least two dogs in total are standing.}
			\end{footnotesize}
		\end{minipage}
	}\\[3pt]
	\fbox{
		\begin{minipage}{0.95\linewidth}
			\centering
			\frame{\colorbox{gray!20}{\frame{\includegraphics[height=16ex]{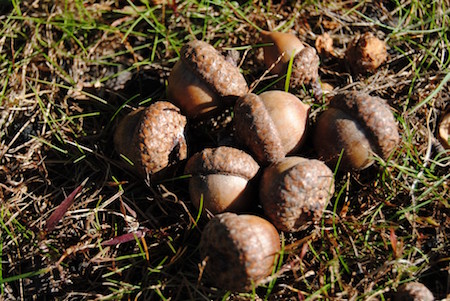}}~{\frame{\includegraphics[height=16ex]{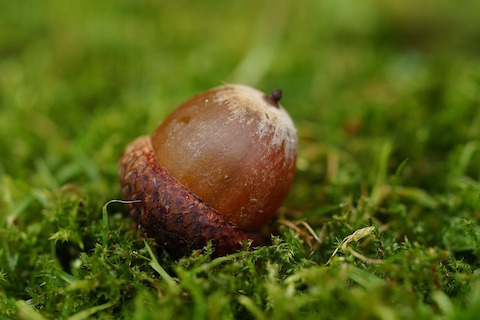}}}}} \\
			\begin{footnotesize}
			\nlstring{One image shows exactly two brown acorns in back-to-back caps on green foliage.}
			\end{footnotesize}
		\end{minipage}
	}
	\caption{Two examples from \realnlvr. 
	Each caption is paired with two images.\footnotemark[2] 
	The task is to predict if the caption is $\truelabel$ or $\falselabel$.
	The examples require addressing challenging semantic phenomena, including resolving \nlstring{twice \dots as} to counting and comparison of objects, and composing cardinality constraints, such as \nlstring{at least two dogs in total} and \nlstring{exactly two}.\footnotemark[3]}
	\label{fig:examples}
\end{figure}

\footnotetext[2]{Appendix~\ref{sec:app:license} contains license information for all photographs used in this paper.}
\footnotetext[3]{The top example is $\truelabel$, while the bottom is $\falselabel$.}
\setcounter{footnote}{3}

We address these limitations with Natural Language Visual Reasoning \emph{for Real} (\realnlvr), a new dataset for reasoning about natural language descriptions of photos. %
The task is to determine if a caption is true with regard to a pair of images. 
Figure~\ref{fig:examples} shows examples from \realnlvr. 
We use images with rich visual content and a data collection process designed to emphasize semantic diversity, compositionality, and visual reasoning challenges. 
Our process reduces the chance of unintentional linguistic biases in the dataset, and therefore the ability of  expressive models to take advantage of them to solve the task. 
Analysis of the data shows that the rich visual input supports diverse language, and that the task requires joint reasoning over the two inputs, including about sets, counts, comparisons, and spatial relations.

Scalable curation of semantically-diverse sentences that describe images requires addressing two key challenges. 
First, we must identify images that are visually diverse enough to support the type of language desired. 
For example, a photo of a single beetle with a uniform background (Table~\ref{tab:interesting_criteria}, bottom left) is likely to elicit only relatively simple sentences about the existence of the beetle and its properties. 
Second, we need a scalable process to collect a large set of captions that demonstrate diverse semantics and  visual reasoning. 

We use a search engine with queries designed to yield sets of similar, visually complex photographs, including of sets of objects and activities, which display real-world scenes. 
We annotate the data through a sequence of crowdsourcing tasks, including filtering for interesting images, writing captions, and validating their truth values. 
To elicit interesting captions, rather than presenting workers with single images, we ask workers for descriptions that compare and contrast four pairs of similar images. 
The description must be $\truelabel$ for two pairs, and $\falselabel$ for the other two pairs. 
Using pairs of images encourages language that composes properties shared between or contrasted among the two images. 
The four pairs are used to create four examples, each comprising an image pair and the description. 
This setup ensures that each sentence appears multiple times with both labels, resulting in a balanced dataset robust to linguistic biases, where a sentence's truth value cannot be determined from the sentence alone, and generalization can be measured using multiple image-pair examples.
This paper includes four main contributions:
(1) a procedure for collecting visually rich images paired with semantically-diverse  language descriptions; 
(2) \realnlvr, which contains $107{,}292$ examples  of captions and image pairs, including $29{,}680$ unique sentences and $127{,}502$ images;
(3) a qualitative linguistically-driven data analysis  showing that our process achieves a broader representation of linguistic phenomena compared to other resources;
and (4) an evaluation with several baselines and state-of-the-art visual reasoning methods on \realnlvr.
The relatively low performance we observe shows that \realnlvr presents a significant challenge, even for methods that perform well on existing visual reasoning tasks. 
\realnlvr is available at \href{http://lil.nlp.cornell.edu/nlvr/}{\tt http://lil.nlp.cornell.edu/nlvr/}.

%% file: related.tex
\section{Related Work and Datasets}
\label{sec:related}

Language understanding in the context of images has been studied within various tasks, including  visual question answering~\cite[e.g.,][]{Zitnick:13abstract,Antol:15vqa}, caption generation~\cite{Chen:16coco}, referring expression resolution~\cite[e.g.,][]{Mitchell:10,Kazemzadeh:14,Mao:16}, visual entailment~\cite{Xie:19}, and binary image selection~\cite{Hu:19bison}. 
Recently, the relatively simple language and reasoning in existing resources motivated datasets that focus on compositional language, 
mostly using synthetic data for language and vision~\cite{Andreas:16nmn,Johnson:16clevr,Kuhnle:17,Kahou:17figureqa,Yang:18cog}.\footnote{A tabular summary of the comparison of \realnlvr to existing resources is available in Table~\ref{tab:dataset_comp}, Appendix~\ref{sec:app:questions}.}
Three exceptions are CLEVR-Humans~\cite{Johnson:17iep}, which includes human-written paraphrases of generated questions for synthetic images; NLVR~\cite{Suhr:17visual-reason}, which uses human-written captions that compare and contrast sets of synthetic images; and GQA~\cite{Hudson:19gqa}, which uses synthetic language grounded in real-world photographs. 
In contrast, we focus on both human-written language and web photographs.
Several methods have been proposed for compositional visual reasoning, including modular neural networks~\cite[e.g.,][]{Andreas:16nmn,Johnson:17iep,Perez:17film,Hu:17n2nmn,Suarez:18,Hu:18,Yao:18cmm,Yi:18} and attention- or memory-based methods~\cite[e.g.,][]{Santoro:17relational,Hudson:18mac,Tan:18cnn-biatt}.
We use \film~\cite{Perez:17film}, \system{N2NMN}~\cite{Hu:17n2nmn}, and \system{MAC}~\cite{Hudson:18mac} for our empirical analysis. 
In our data, we use each sentence in multiple examples, but with different labels. 
This is related to recent visual question answering datasets that aim to require models to consider both image and question to perform well~\cite{Zhang:16,Goyal:17,Li:17,Agrawal:17,Agrawal:18}.
Our approach is inspired by the collection of NLVR, where workers were shown a set of similar images and asked to write a sentence $\truelabel$ for some images, but $\falselabel$ for the others~\cite{Suhr:17visual-reason}.
We adapt this method to web photos, including introducing a process to identify images that support complex reasoning and designing incentives for the more challenging writing task.

%% file: collection.tex
\section{Data Collection}
\label{sec:data}

\input{collection_fig}
Each example  in \realnlvr includes a pair of images and a natural language sentence.
The task is to determine whether the sentence is $\truelabel$ or $\falselabel$ about the pair of images. 
Our goal is to collect a large corpus of grounded semantically-rich descriptions that require diverse types of reasoning, including about sets, counts, and comparisons.
We design a process to identify images that enable such types of reasoning, collect grounded natural language descriptions, and label them as $\truelabel$ or $\falselabel$.
While we use image pairs, we do not explicitly set the task of describing the differences between the images or identifying which image matches the sentence better~\cite{Hu:19bison}. 
We use pairs to enable comparisons and set reasoning between the objects that appear in the two images. 
Figure~\ref{fig:collection} illustrates our data collection procedure. 
For further discussion on the design decisions for our task and data collection implementation, please see appendices~\ref{sec:app:questions} and~\ref{sec:app:data_collection}.

\input{query_heuristics}

\begin{table}[t]
\centering
\begin{footnotesize}
\begin{tabular}{|m{2.7cm}m{4cm}|} \hline
\multicolumn{2}{|c|}{\textbf{Positive Examples and Criteria}} \\ \hline
\vspace{2pt}\frame{\includegraphics[width=1.\linewidth,clip,trim=0 0 0 70]{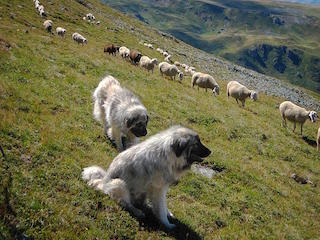}}\vspace{-0.8pt} & Contains more than one instance of the synset. \\ 
\vspace{2pt}\frame{\includegraphics[width=1.\linewidth,clip,trim=0 100 0 150]{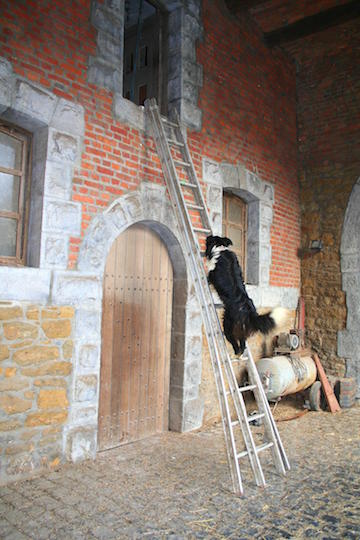}}\vspace{-0.8pt} & Shows an instance of the synset interacting with other objects.\\ 
\vspace{2pt}\frame{\includegraphics[width=1.\linewidth]{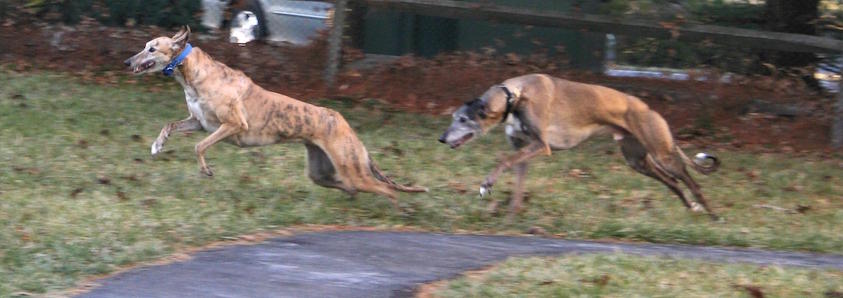}}\vspace{-0.8pt} & Shows an instance of the synset performing an activity.\\ 
\vspace{2pt}\frame{\includegraphics[width=1.\linewidth]{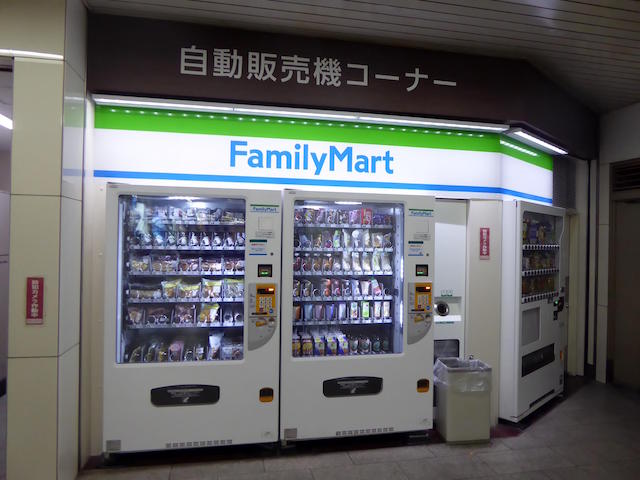}}\vspace{-0.8pt} & Displays a set of diverse objects or features. \\\hline\hline
\multicolumn{2}{|c|}{\textbf{Negative Examples}} \\ \hline
& \\[-0.8em]
\multicolumn{2}{|c|}{\frame{\includegraphics[height=0.165\linewidth]{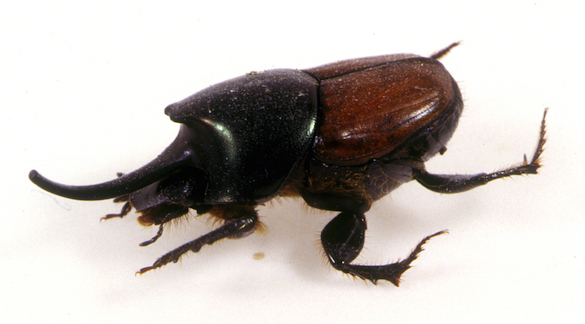}} \frame{\includegraphics[height=0.165\linewidth]{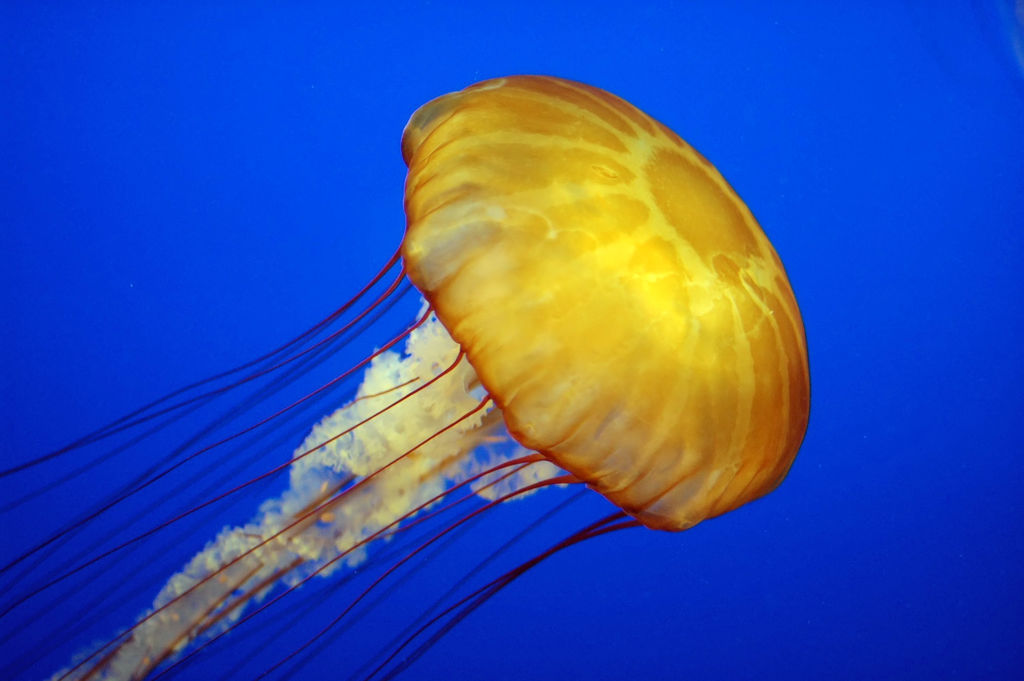}} \frame{\includegraphics[height=0.165\linewidth]{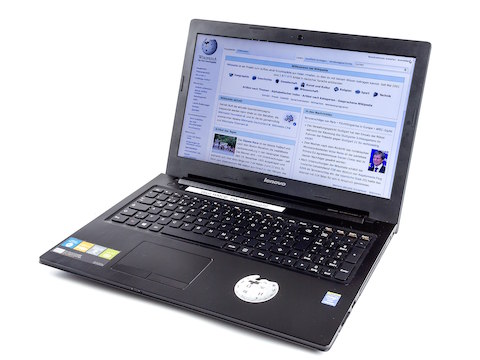}} \frame{\includegraphics[height=0.165\linewidth]{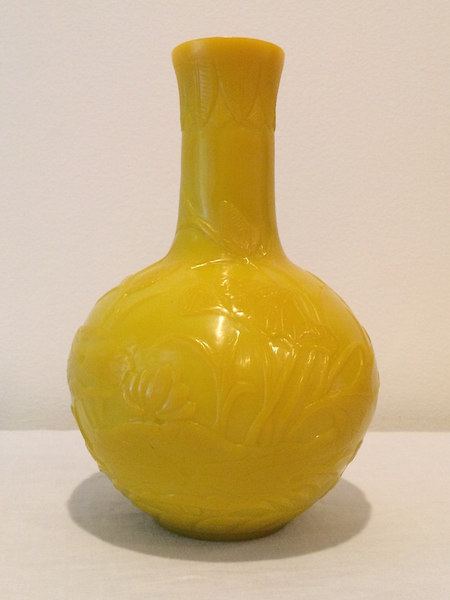}}} \\ \hline
\end{tabular}
\end{footnotesize}
\caption{Positive and negative examples of interesting images.}
\label{tab:interesting_criteria}
\end{table}

\subsection{Image Collection}\label{sec:image_collection}

We require sets of images where the images in each set are detailed but similar enough such that comparison will require use of a diverse set of reasoning skills, more than just object or property identification. 
Because existing image resources, such as ImageNet~\cite{Russakovsky:15} or COCO~\cite{Lin:14coco}, do not provide such grouping and mostly include relatively simple object-focused scenes, we collect a new set of images. 
We retrieve sets of images with similar content using search queries generated from synsets from the ILSVRC2014 ImageNet challenge~\cite{Russakovsky:15}. 
This  correspondence to ImageNet synsets  allows researchers to use pre-trained image featurization  models, and focuses the challenges of the task not on object detection, but compositional reasoning challenges.

\paragraph{ImageNet Synsets Correspondence} 
We identify a subset of the $1{,}000$ synsets in ILSVRC2014 that often appear in rich contexts.
For example, an \synset{acorn} often appears in images with other \synset{acorns}, while a \synset{seawall} almost always appears alone.
For each synset, we issue five queries to the Google Images search engine\footnote{\href{https://images.google.com/}{\tt https://images.google.com/}} using query expansion heuristics. 
The heuristics are designed to retrieve images that support complex reasoning, including images with groups of entities, rich environments, or entities participating in activities. 
For example, the expansions for the synset \texttt{acorn} will include \texttt{two acorns} and \texttt{acorn fruit}.   
The heuristics are specified in Table~\ref{tab:query_heuristics}. 
For each query, we use the Google similar images tool for each of the first five images to retrieve the seven non-duplicate most similar images. 
This results in five sets of eight similar images per query,\footnote{At the time of publication, the similar images tool is available at the ``View more'' link in the list of related images after expanding the results for each image. Images are ranked by similarity, where more similar images appear higher.} $25$ sets in total.   
If at least half of the images in a set were labeled as interesting according to the criteria in Table~\ref{tab:interesting_criteria}, %
the synset is awarded one point. 
We choose the $124$ synsets with the most points.\footnote{We pick $125$ and remove one set due to high image pruning rate in later stages.} 
The $124$ synsets are distributed evenly among animals and objects.
This annotation was performed by the first two authors and student volunteers,  is only used for identifying synsets, and is separate from the image search  described below.

\paragraph{Image Search}
We use the Google Images search engine to find sets of similar images (Figure~\ref{fig:collection}a). 
We apply the query generation heuristics to the $124$ synsets. 
We use all synonyms in each synset~\cite{Deng:14,Russakovsky:15}.
For example, for the synset \synset{timber wolf}, we use the synonym set $\{$\synset{timber wolf}, \synset{grey wolf}, \synset{gray wolf}, \synset{canis lupus}$\}$.
For each generated query, we download sets containing at most $16$ related images.

\paragraph{Image Pruning}
We use two crowdsourcing tasks to (1) prune the sets of images, and (2) construct sets of eight images to use in the sentence-writing phase.
In the first task, we remove low-quality images from each downloaded set of similar images (Figure~\ref{fig:collection}b).
We display the image set and the synset name, and ask a worker to remove any images that do not load correctly; images that contain inappropriate content, non-realistic artwork, or collages; or images that do not contain an instance of the corresponding synset. 
This results in sets of sixteen or fewer similar images.
We discard all sets with fewer than eight images. 

The second task further prunes these sets by removing duplicates and down-ranking non-interesting images (Figure~\ref{fig:collection}c).
The goal of this stage is to collect sets that contain enough interesting images.  
Workers are asked to remove duplicate images, and mark images that are not \emph{interesting}.
An image is interesting if it fits any of the criteria in Table~\ref{tab:interesting_criteria}.
We ask workers not to mark an image if they consider it interesting for any other reason.
We discard  sets with fewer than three  interesting images. 
We sort the images in descending order according to first interestingness, and second similarity, and keep the top eight. 
\subsection{Sentence Writing}
\label{sec:sentence_writing}
\input{examples}

Each set of eight images is used for a sentence-writing task. 
We randomly split the set into four pairs of images. 
Using pairs encourages comparison and set reasoning within the pairs. 
Workers are asked to select two of the four pairs and write a sentence that is  $\truelabel$ for the selected pairs, but $\falselabel$ for the unselected pairs.
Allowing workers to select pairs themselves makes the sentence-writing task easier  than with random selection, which may create tasks that are impossible to complete.
Writing requires finding similarities and differences between the pairs, which encourages  compositional language~\cite{Suhr:17visual-reason}.

In contrast to the collection process of NLVR, using real images does not allow for as much control over their content, in some cases permitting workers to write simple sentences.
For example, a worker could write a sentence stating the existence of a single object if it was only present in both selected pairs, which  is avoided in NLVR by controlling for the objects in the images.
Instead, we define more specific guidelines for the workers for writing sentences, including asking to avoid subjective opinions, discussion of properties of photograph, mentions of text, and simple object identification. 
We include more details and examples of these guidelines in Appendix~\ref{sec:app:data_collection}. 

\subsection{Validation}\label{sec:validation}
We split each sentence-writing task into four examples, where the sentence is paired with each pair of images. 
Validation  ensures that the selection of each image pair reflects its truth value.
We show each example independently to a worker, and ask them to label it as $\truelabel$ or $\falselabel$.
The worker may also report the sentence as nonsensical.
We keep all non-reported examples where the validation label is the same as the initial label indicated by the sentence-writer's selection.
For example, if the image pair is initially selected during sentence-writing, the sentence-writer intends the sentence to be $\truelabel$ for the pair, so if the validation label is $\falselabel$, this example is removed.

\subsection{Splitting the Dataset}\label{sec:split}
We assign a random $20\%$ of the examples passing validation to development and testing, ensuring that examples from the same initial  set of eight images do not appear across the split.
For these examples, we collect four additional validation judgments to estimate agreement and human performance. 
We remove from this set examples  where two or more of the extra judgments disagreed with the existing label (Section~\ref{sec:validation}).
Finally, we create equal-sized splits for a development set and two test sets, ensuring that original image sets do not appear in multiple splits of the data (Table~\ref{tab:sizes}).

\subsection{Data Collection Management}\label{sec:data:management}

We use a tiered system with bonuses to encourage workers to write linguistically diverse sentences. After every round of annotation, we sample examples for each worker and give bonuses to workers that follow our writing guidelines well. Once workers perform at a sufficient level, we allow them access to a larger pool of tasks. 
We also use qualification tasks to train workers.
The mean cost per unique sentence in our dataset is \$0.65; the mean cost per example is \$0.18. 
Appendix~\ref{sec:app:data_collection} provides additional details about our bonus system, qualification tasks, and costs. 

\subsection{Collection Statistics}\label{sec:data:stat}

We collect $27{,}678$ sets of related images and a total of $387{,}426$ images (Section~\ref{sec:image_collection}).
Pruning low-quality images leaves $19{,}500$ sets and $250{,}862$ images. 
Most images are removed for not containing an instance of the corresponding synset or for being non-realistic artwork or a collage of images. 
We construct  $17{,}685$ sets of eight images each. 

We crowdsource $31{,}418$ sentences (Section~\ref{sec:sentence_writing}).
We create two writing tasks for each set of eight images. 
Workers may flag sets of images if they should have been removed in earlier stages; for example, if they contain duplicate images.
Sentence-writing tasks that remain without annotation after three days are removed.

During validation, $1{,}875$ sentences are reported as nonsensical.
$108{,}516$ examples pass validation; i.e., the validation label matches the initial selection for the pair of images (Section~\ref{sec:validation}).
Removing low-agreement examples in the development and test sets yields a dataset of $107{,}292$ examples, $127{,}502$ unique images, and $29{,}680$ unique sentences. 
Each unique sentence is paired with an average of $3.6$ pairs of images. 
Table~\ref{tab:paired_examples} shows examples of three unique sentences from \realnlvr.
Table~\ref{tab:sizes} shows the sizes of the data splits, including train, development, a public test set (Test-P),  and an unreleased test set (Test-U).

\begin{table}[t]
\begin{footnotesize}
\begin{center}
\begin{tabular}{|p{2.5cm}|c|c|} \hline
& \textbf{Unique sentences} & \textbf{Examples}  \\ \hline
Train & $23{,}671$ & $86{,}373$  \\ \dline
Development & $2{,}018$ & $6{,}982$  \\ \dline
Test-P & $1{,}995$ & $6{,}967$  \\ \dline
Test-U & $1{,}996$ & $6{,}970$ \\ \hline
Total & $29{,}680$ & $107{,}292$  \\ \hline
\end{tabular}
\end{center}
\end{footnotesize}
\caption{\realnlvr data splits.}
\label{tab:sizes}
\end{table}

%% file: collection_fig.tex
\newcommand{\subfigwidth}{0.82\linewidth}

\begin{figure*}[ht!]
\begin{footnotesize}
(a) \textbf{Find Sets of Images:} 
The query \synset{two acorns} is issued to the search engine.
The leftmost image appears in the list of results.
The Similar Images tool is used to find a set of images, shown on the right,  similar to this image.
\begin{center}
		\begin{minipage}{0.95\linewidth}
			\centering
			\frame{\includegraphics[trim={247 553 290 8},clip,width=\subfigwidth]{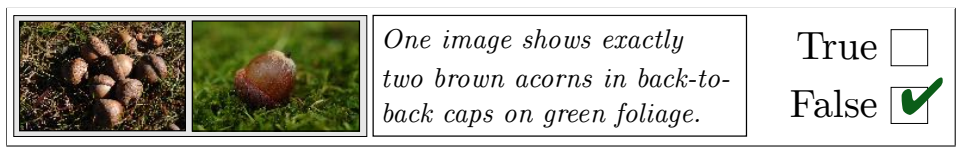}}
		\end{minipage}
\end{center}

(b) \textbf{Image Pruning:} Crowdworkers are given the synset name and identify low-quality images to be removed. In this example, one image is removed because it does not show an instance of the synset \synset{acorn}. 

\begin{center}
\begin{minipage}{0.95\linewidth}
\centering
\frame{\includegraphics[trim={240 485 241 102},clip,width=\subfigwidth]{figs/collection_tall.pdf}}
\end{minipage}
\end{center}

(c) \textbf{Set Construction:} Crowdworkers decide whether each of the remaining images is interesting. 
In this example, three images are marked as non-interesting (top row) because they contain only a single instance of the synset.
The images are re-ordered (bottom row) so that interesting images appear before non-interesting images, and the top eight images are used to form the set.
In this example, the set is formed using the leftmost eight images.

\begin{center}
\begin{minipage}{0.95\linewidth}
\centering
\frame{\includegraphics[trim={241 453 274 132},clip,width=\subfigwidth]{figs/collection_tall.pdf}}
\end{minipage}
\end{center}

\begin{center}
\begin{minipage}{0.95\linewidth}
\centering
\frame{\includegraphics[trim={241 428 242 159},clip,width=\subfigwidth]{figs/collection_tall.pdf}}
\end{minipage}
\end{center}

(d) \textbf{Sentence Writing:} The images in the set are randomly paired and shown to the worker.
The worker selects two pairs, and writes a sentence that is $\truelabel$ for the two selected pairs but $\falselabel$ for the other two pairs.

\begin{center}
\begin{minipage}{0.95\linewidth}
\centering
\frame{\includegraphics[trim={241 369.5 281 190},clip,width=\subfigwidth]{figs/collection_tall.pdf}}
\end{minipage}
\end{center}

(e) \textbf{Validation:} Each pair forms an example with the written sentence. 
Each example is shown to a worker to re-label.

\begin{center}
\begin{minipage}{0.95\linewidth}
\centering
\frame{\includegraphics[trim={262 295 289 282},clip,width=\subfigwidth]{figs/collection_tall.pdf}}
\end{minipage}
\end{center}

\end{footnotesize}
\caption{Diagram of the data collection process, showing how a single example from the training set is constructed. 
Steps (a)--(c) are described in Section~\ref{sec:image_collection}; step (d) in Section~\ref{sec:sentence_writing}; and step (e) in Section~\ref{sec:validation}.}
\label{fig:collection}
\end{figure*}

%% file: query_heuristics.tex
\begin{table*}[t]
\begin{footnotesize}
\centering
\begin{tabular}{|l|p{4.4cm}|p{8.4cm}|} \hline
\textbf{Heuristic} & \textbf{Examples} \newline (\synset{synset synonym} $\rightarrow$ \synset{query}) & \textbf{Description} \\ \hline
Quantities & \synset{cup} $\rightarrow$ \synset{group of cups} & Add numerical phrases or manually-identified collective nouns to the synonym. These queries result in images containing multiple examples of the synset. \\ \dline
Hypernyms & \synset{flute} $\rightarrow$ \synset{flute woodwind} & Add direct or indirect hypernyms from WordNet~\cite{Miller:92}. Applied only to the non-animal synsets. This heuristic increases the diversity of images retrieved for the synset~\cite{Deng:14}.\\ \dline
Similar words & \synset{banana} $\rightarrow$ \synset{banana pear} & Add concrete nouns whose cosine similarity with the synonym is greater than $0.35$ in the embedding space of Google News word2vec embeddings~\cite{Mikolov:13word2vec}. Applied only to non-animal synsets. These queries result in images containing a variety of different but related object types. \\ \dline
Activities & \synset{beagle} $\rightarrow$ \synset{beagles eating} & Add manually-identified verbs describing common activities of animal synsets. Applied only to animal synsets. This heuristic results in images of animals participating in activities, which encourages captions with a diversity of entity properties. \\ \hline
\end{tabular}
\end{footnotesize}
\caption{The four heuristics used to generate search queries from synsets.}
\label{tab:query_heuristics}
\end{table*}

%% file: examples.tex
\newcommand{\imgpairmain}[2]{\centering\vspace{0.4em}\frame{\colorbox{gray!20}{\frame{\includegraphics[height=10ex]{{#1}}}~{\frame{\includegraphics[height=10ex]{{#2}}}}}}}
\newcommand{\labcolwidth}{0.5cm}
\newcommand{\colwidth}{4.4cm}

\begin{table*}
\centering\footnotesize
\setlength{\tabcolsep}{0.1cm}
\begin{tabular}{m{0.8cm}|m{4.51cm}|c|m{4.76cm}|c|m{4.6cm}|m{0cm}} \cline{2-2} \cline{4-4} \cline{6-6}
   $\truelabel$ &  \imgpairmain{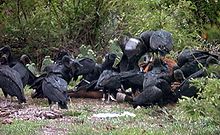}{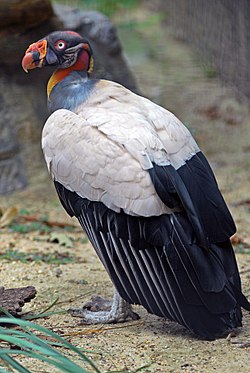} &&\imgpairmain{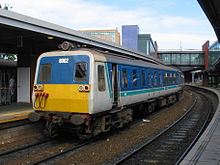}{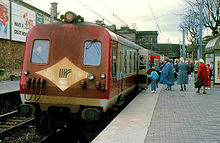}&&  \imgpairmain{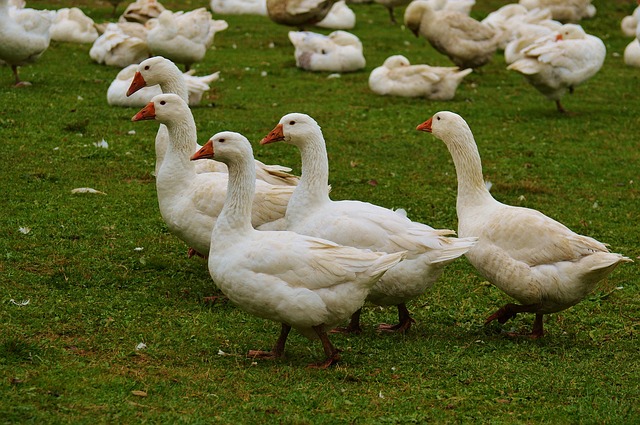}{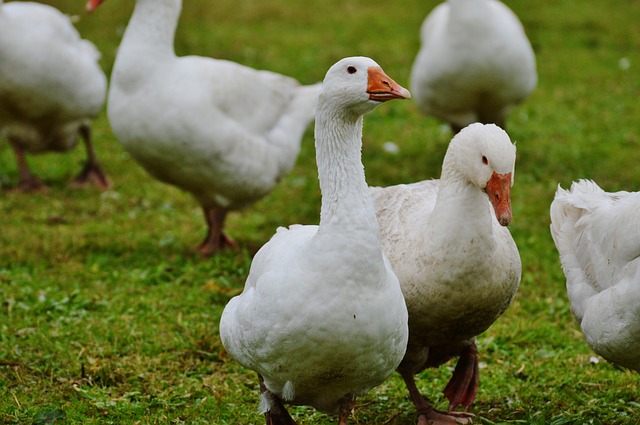} & \\
   $\falselabel$  & \imgpairmain{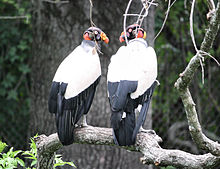}{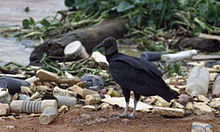} && \imgpairmain{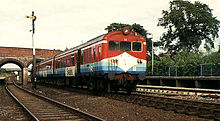}{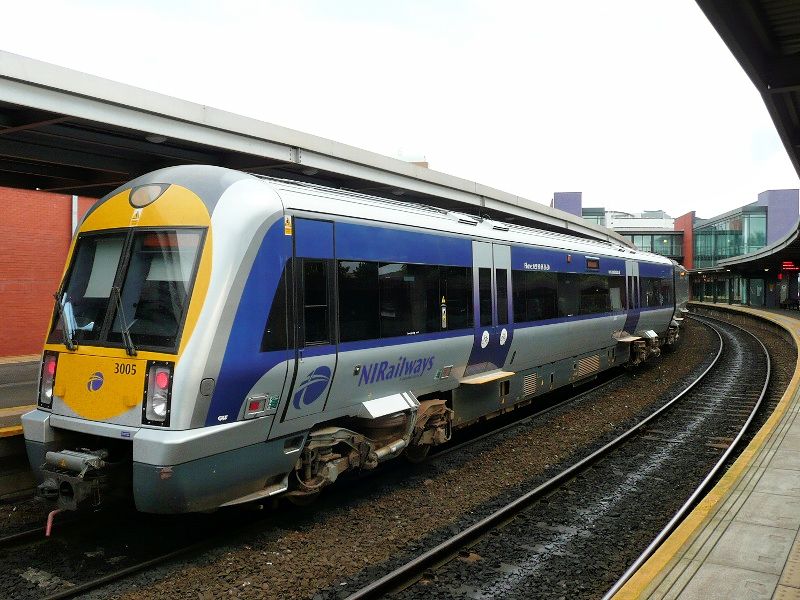}  && \imgpairmain{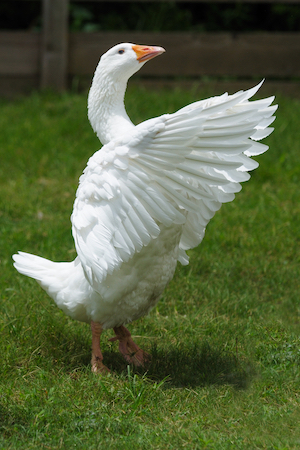}{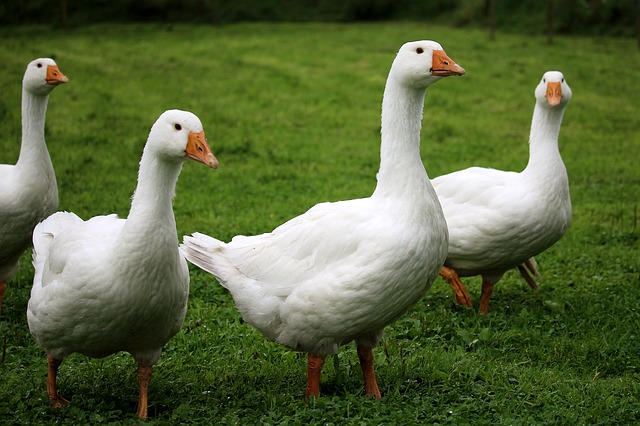} & \\
   & \nlstring{One image contains a single vulture in a standing pose with its head and body facing leftward, and the other image contains a group of at least eight vultures.} && \nlstring{There are two trains in total traveling in the same direction.}&&  \nlstring{There are more birds in the image on the left than in the image on the right.} & \\ \cline{2-2} \cline{4-4} \cline{6-6}
\end{tabular}
\caption{Six examples with three different sentences from \realnlvr. For each sentence, we show two examples using different image-pairs, each with a different label.}
    \label{tab:paired_examples}
\end{table*}

%% file: analysis.tex
\section{Data Analysis}
\label{sec:analysis}

We perform quantitative and qualitative analysis using the training and development sets.

\paragraph{Agreement}
Following validation, $8.5\%$ of  the examples  not reported during validation are removed due to  disagreement between the validator's label and the initial selection of the image pair~(Section~\ref{sec:validation}).\footnote{The validator is the same worker as the sentence-writer for $11.5\%$ of examples. In these cases, the validator agrees with themselves $96.7\%$ of the time. For examples where the sentence-writer and validator were not the same person, they agree in $90.8\%$ of examples.}
We use the five validation labels we collect for the development and test sets to compute   Krippendorff's $\alpha$ and Fleiss' $\kappa$ to measure agreement~\cite{Cocos:15,Suhr:17visual-reason}. 
Before removing low-agreement examples (Section~\ref{sec:split}), $\alpha=0.906$ and $\kappa=0.814$.
After removal,  $\alpha=0.912$ and $\kappa=0.889$, indicating almost perfect agreement~\cite{Landis:77}.

\paragraph{Synsets}

Each synset is associated with $\mu = 752.9 \pm 205.7$ examples. 
The five most common synsets are \synset{gorilla}, \synset{bookcase}, \synset{bookshop}, \synset{pug}, and \synset{water buffalo}.
The five least common synsets are \synset{orange}, \synset{acorn}, \synset{ox}, \synset{dining table}, and \synset{skunk}.
Synsets appear in equal proportions across the four splits.

\input{lengths}

\begin{table*}[t]
\begin{footnotesize}
\begin{center}
\begin{tabular}{|p{2.3cm}|c|c|c|c||p{7.3cm}|} 
	\hline
    & \textbf{VQA}& \textbf{GQA} & \textbf{NLVR} &\textbf{\realnlvr} & \textbf{Example from \realnlvr} \\ 
	&  \textbf{(real)} \% & \% & \% & \% & \\ 
	\hline
	\hline
	\multicolumn{5}{|l|}{Semantics}  \\
	\hline
	\multirow{2}{*}{Cardinality (hard)} &  \multirow{2}{*}{$11.5$} & \multirow{2}{*}{$0$}& \multirow{2}{*}{$66$} & \multirow{2}{*}{$41.1$} & \nlstring{\textbf{Six} rolls of paper towels are enclosed in a plastic package with the brand name on it.} \\ \dline
	{Cardinality (soft)} & {$1$} & {$0$} & {$16$} & {$23.6$} & \nlstring{\textbf{No more than two} cheetahs are present.}  \\ \dline
	Existential & $11.5$ & $16.5$ & $88$ & $23.6$ & \nlstring{\textbf{There are} at most 3 water buffalos in the image pair.}  \\ \dline
		\multirow{2}{*}{Universal} & \multirow{2}{*}{$1$} & \multirow{2}{*}{$4.5$} & \multirow{2}{*}{$7.5$} & \multirow{2}{*}{$16.8$} & \nlstring{In one image there is a line of fence posts with one large darkly colored bird on top of \textbf{each} post.}  \\ \dline
		\multirow{2}{*}{Coordination} & \multirow{2}{*}{$5$} & \multirow{2}{*}{$21.5$} & \multirow{2}{*}{$17$} & \multirow{2}{*}{$33.3$} & \nlstring{Each image contains only one wolf, \textbf{and} all images include snowy backdrops.}  \\  \dline 
		\multirow{2}{*}{Coreference} & \multirow{2}{*}{$6.5$} & \multirow{2}{*}{$0.5$} & \multirow{2}{*}{$3$} & \multirow{2}{*}{$14.6$} & \nlstring{there are four or more animals very close to \textbf{each other} on the grass in the image to the left.} \\  \dline
		Spatial Relations & $42.5$ & $43$ & $66$ & $49$ & \nlstring{A stylus is \textbf{near} a laptop in one of the images.} \\ \dline
		\multirow{2}{*}{Comparative} & \multirow{2}{*}{$1$} & \multirow{2}{*}{$2$} & \multirow{2}{*}{$3$} & \multirow{2}{*}{$8$} & \nlstring{There are \textbf{more} birds in the image on the right \textbf{than} in the image on the left.}  \\ \dline
		Presupposition  & $80$ & $79$ & $19.5$ & $20.6$ & \nlstring{A cookie sits in \textbf{the dessert} in the image on the left.}\\  \dline
	    \multirow{2}{*}{Negation} & \multirow{2}{*}{$1$} & \multirow{2}{*}{$2.5$ }& \multirow{2}{*}{$9.5$} & \multirow{2}{*}{$9.6$}  & \nlstring{The front paws of the dog in the image on the left are \textbf{not touching} the ground.}\\
	\hline
	\hline
	\multicolumn{5}{|l|}{Syntactic Ambiguity} \\
	\hline
	\multirow{3}{*}{CC Attachment} & \multirow{3}{*}{$0$} & \multirow{3}{*}{$2.5$} & \multirow{3}{*}{$4.5$} & \multirow{3}{*}{$3.8$} &  \nlstring{The left image shows a cream-layered dessert in a footed clear glass which includes sliced peanut butter cups \textbf{and} brownie chunks.} \\ \dline
	\multirow{2}{*}{PP Attachment} & \multirow{2}{*}{$3$} & \multirow{2}{*}{$6.5$} & \multirow{2}{*}{$23$} & \multirow{2}{*}{$11.5$} & \nlstring{At least one panda is sitting near a fallen branch \textbf{on the ground}.}\\ \dline
	SBAR \newline Attachment & \multirow{2}{*}{$0$} & \multirow{2}{*}{$5$} & \multirow{2}{*}{$2$} & \multirow{2}{*}{$1.9$} & \nlstring{Balloons float in a blue sky with dappled clouds on strings \textbf{that} angle rightward, in the right image.} \\
	\hline
\end{tabular}
\end{center}
\end{footnotesize}
\caption{Linguistic analysis of sentences from \realnlvr, GQA, VQA, and NLVR. 
We analyze $800$ development sentences from \realnlvr and $200$ from each of the other datasets for the presence of  semantic and syntactic phenomena described in \citet{Suhr:17visual-reason}.
We report the proportion of examples containing each phenomenon.}
\label{tab:cats}
\end{table*}

\paragraph{Language}
\realnlvr's vocabulary contains $7{,}457$ word types, significantly larger than NLVR, which has $262$ word types.
Sentences in \realnlvr are on average $14.8$ tokens long, whereas NLVR has a mean sentence length of $11.2$. 
Figure~\ref{fig:lengths} shows the distribution of sentence lengths compared to related corpora.
\realnlvr shows a similar distribution to NLVR, but with a longer tail. 
\realnlvr contains longer sentences than the questions of VQA~\cite{Antol:15vqa}, GQA~\cite{Hudson:19gqa}, and CLEVR-Humans~\cite{Johnson:17iep}.
Its distribution is similar to MSCOCO~\cite{Chen:15coco}, which also contains captions, and CLEVR~\cite{Johnson:16clevr}, where the language is synthetically generated.

We analyze $800$ sentences from the development set for occurrences of semantic and syntactic phenomena (Table~\ref{tab:cats}).
We compare with the  $200$-example analysis of VQA and NLVR from \citet{Suhr:17visual-reason}, and $200$ examples from the balanced split of GQA.
Generally, \realnlvr has similar linguistic diversity to NLVR, showing broader representation of linguistic phenomena than VQA and GQA. 
One noticeable difference from NLVR is less use of hard cardinality. 
This is possibly due to how NLVR is designed to use a very limited set of object attributes, which encourages  writers  to rely on accurate counting for discrimination more often. 
We include further analysis in Appendix~\ref{sec:app:analysis}. 

%% file: lengths.tex
\definecolor{orange}{RGB}{229,158,54}
\definecolor{sky}{RGB}{96,181,226}
\definecolor{bluegreen}{RGB}{26,158,116}
\definecolor{yellow}{RGB}{240,227,84}
\definecolor{blueblue}{RGB}{21,116,177}
\definecolor{vermillion}{RGB}{234,21,24}
\definecolor{redpurple}{RGB}{202,122,166}
\definecolor{purplepurple}{RGB}{206, 72, 226}

\begin{figure}[t]
\begin{center}
\begin{footnotesize}
\begin{tikzpicture}
 \begin{axis}[
    width=1.05\columnwidth,
        height=0.53\columnwidth,
        xmin=1,xmax=40,
        ymin=0, ymax=30,
        xtick={5,10,15,20,25,30, 35, 40},
        ytick={0,5,10,15,20,25,30},
        bar width=12pt,
        xlabel style={yshift=1ex,},
        ylabel style={yshift=-2.5ex,},
        xlabel=Sentence length,
        ylabel=\% of sentences,
        legend style={font=\scriptsize,at={(0.5,1.05)},anchor=south},
        legend columns=4,
        legend cell align={left},
        legend transposed=true]
        
     \addplot[smooth,color=sky,style={thick}] coordinates {
		(1,0)
		(2,0)
		(3,0)
		(4,2.97)
		(5,14.43)
		(6,26.24)
		(7,19.99)
		(8,15.91)
		(9,9.85)
		(10,4.8)
		(11,2.6)
		(12,1.46)
		(13,0.77)
		(14,0.4)
		(15,0.23)
		(16,0.14)
		(17,0.08)
		(18,0.05)
		(19,0.03)
		(20,0.02)
		(21,0.01)
		(22,0)
		(23,0)
		(24,0)
		(25,0)
		(26,0)
	(27,0)
	(28,0)
	(29,0)
	(30,0)
	(31,0)
	(32,0)
	(33,0)
	(34,0)
	(35,0)
	(36,0)
	(37,0)
         (38,0)
         (39,0)
         (40,0)
         (41,0)
     }; 
     \addlegendentry{VQA (real)}   
     
           \addplot[smooth,densely dotted,color=sky,style={thick}] coordinates {
		(1,0)
		(2,0)
		(3,0)
		(4,5.83)
		(5,15.18)
		(6,26.8)
		(7,17.68)
		(8,15.52)
		(9,9.03)
		(10,4.53)
		(11,2.5)
		(12,1.38)
		(13,0.68)
		(14,0.39)
		(15,0.21)
		(16,0.11)
		(17,0.08)
		(18,0.04)
		(19,0.02)
		(20,0.02)
		(21,0.01)
				(22,0)
		(23,0)
		(24,0)
		(25,0)
		(26,0)
	(27,0)
	(28,0)
	(29,0)
	(30,0)
	(31,0)
	(32,0)
	(33,0)
	(34,0)
	(35,0)
	(36,0)
	(37,0)
         (38,0)
         (39,0)
         (40,0)
         (41,0)	
     }; 
     \addlegendentry{VQA (abstract)}
     
           \addplot[densely dotted,smooth,color=vermillion,style=thick] coordinates {
         (1,0)
         (2,0)
         (3,0)
         (4,0.03)
         (5,0.38)
         (6,1.8)
         (7,2.3)
         (8,1.38)
         (9,1.24)
         (10,1.37)
         (11,2.57)
         (12,4.45)
         (13,6.93)
         (14,8.72)
         (15,8.18)
         (16,5.72)
         (17,4.24)
         (18,4.31)
         (19,5.09)
         (20,5.43)
         (21,4.62)
         (22,3.98)
         (23,3.58)
         (24,3.36)
         (25,3.14)
         (26,2.87)
         (27,2.57)
         (28,2.35)
         (29,2.03)
         (30,1.81)
         (31,1.5)
         (32,1.24)
         (33,0.97)
         (34,0.71)
         (35,0.49)
         (36,0.31)
         (37,0.19)
         (38,0.1)
         (39,0.05)
         (40,0.02)
         (41,0.01)
	};
	\addlegendentry{CLEVR}
	
	        \addplot[densely dotted,smooth,color=purplepurple,style={thick}] coordinates{
(1,0)
(2,0)
(3,0)
(4,0.05)
(5,2.13)
(6,14.03)
(7,21.29)
(8,12.02)
(9,10.49)
(10,8.80)
(11,6.43)
(12,6.05)
(13,4.78)
(14,3.87)
(15,2.82)
(16,2.05)
(17,1.51)
(18,0.89)
(19,0.71)
(20,0.61)
(21,0.37)
(22,0.29)
(23,0.26)
(24,0.15)
(25,0.12)
(26,0.06)
(27,0.06)
(28,0.06)
(29,0.04)
(30,0.02)
(31,0.00)
(32,0.01)
(33,0.01)
(34,0)
(35,0.00)
(36,0)
(37,0)
(38,0)
(39,0)
(40,0)
(41,0)
   };
   \addlegendentry{CLEVR-Humans}

           \addplot[smooth,color=bluegreen,style=thick] coordinates {
         (1,0)
         (2,0)
         (3,0)
         (4,0)
         (5,0)
         (6,0)
         (7,0.39)
         (8,5.03)
         (9,17.34)
         (10,20.85)
         (11,19.34)
         (12,13.91)
         (13,9.07)
         (14,5.40)
         (15,3.13)
         (16,1.86)
         (17,1.17)
         (18,0.74)
         (19,0.49)
         (20,0.35)
         (21,0.24)
         (22,0.16)
         (23,0.13)
         (24,0.09)
         (25,0.07)
         (26,0.05)
         (27,0.04)
         (28,0.03)
         (30,0.02)
         (31,0.01)
         (32,0.01)
         (33,0.01)
         (34,0.01)
         (35,0.01)
         (36,0.01)
         (37,0.01)
         (38,0)
         (39,0)
         (40,0)
         (41,0)
	};
	\addlegendentry{MSCOCO}

	\addplot[smooth,style={thick},color=purplepurple] 
	coordinates {
	(	4,	0.8206968528)
(5,	5.372434334)
(6,	13.08891952)
(7,	13.60209923)
(8,	10.8723962)
(9,	7.790248376)
(10,	12.05976958)
(11,	8.709172122)
(12,	7.519845367)
(13,	4.409978215)
(14,	3.864056213)
(15,	3.339714361)
(16,	2.709239095)
(17,	1.781385632)
(18,	1.301878403)
(19,	1.070170837)
(20,	0.7321438205)
(21,	0.388535731)
(22,	0.3406315171)
(23,	0.1720830985)
(24,	0.03934656792)
(25,	0.01013895013)
(26,	0.00353467986)
(27,	0.0008371610196)
(28,	0.0005581073464)
(29,	0.0001860357821)
	};
	\addlegendentry{GQA}

              \addplot[densely dotted,smooth,style={thick}] coordinates {
	(1,0)
	(2,0)
	(3,0)
	(4,0.05)
	(5,6.17)
	(6,1.20)
	(7,1.8)
	(8,8.98)
	(9,13.76)
	(10,13.2)
	(11,9.7)
	(12,12.22)
	(13,11.67)
	(14,4.81)
	(15,6.96)
	(16,2.91)
	(17,3.21)
	(18,1.22)
	(19,0.76)
	(20,0.51)
	(21,0.46)
	(22,0.21)
	(23,0.05)
	(24,0.05)
	(25,0.08)
	(26,0)
	(27,0)
	(28,0)
	(29,0)
	(30,0)
	(31,0)
	(32,0)
	(33,0)
	(34,0)
	(35,0)
	(36,0)
	(37,0)
         (38,0)
         (39,0)
         (40,0)
         (41,0)
              };     
     \addlegendentry{NLVR}

                   \addplot[smooth,style={ultra thick}] coordinates {
(1,0)
(2,0)
(3,0)
(4,0.41)
(5,1.16)
(6,2.11)
(7,4.22)
(8,5.51)
(9,6.45)
(10,7.11)
(11,8.17)
(12,8.67)
(13,7.94)
(14,7.38)
(15,5.93)
(16,4.89)
(17,4.00)
(18,3.45)
(19,2.99)
(20,2.57)
(21,2.35)
(22,1.96)
(23,1.64)
(24,1.63)
(25,1.57)
(26,1.31)
(27,1.03)
(28,0.88)
(29,0.94)
(30,0.69)
(31,0.59)
(32,0.54)
(33,0.52)
(34,0.34)
(35,0.22)
(36,0.22)
(37,0.16)
(38,0.11)
(39,0.1)
(40,0.08)
(41,0.06)

              };     
     \addlegendentry{\realnlvr}
    
    \end{axis}
\end{tikzpicture}
\end{footnotesize}
\end{center}
\caption{Distribution of sentence lengths. Dotted curves represent datasets with synthetic images.}
\label{fig:lengths}
\end{figure}
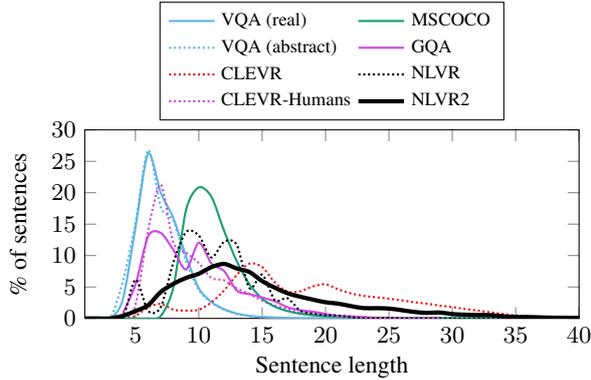  

%% file: human_performance.tex
\section{Estimating Human Performance}
\label{sec:human_performance}

We use the additional labels of the development and test examples to estimate human performance.
We group these labels according to workers.  
We do not consider cases where the worker labels a sentence written by themselves. 
For each worker, we measure their performance as the proportion of their judgements that matches the gold-standard label, which is the original validation label. 
We compute the average and standard deviation performance over workers with at least $100$ such additional validation judgments, a total of $68$ unique workers.
Before pruning low-agreement examples (Section~\ref{sec:split}), the average performance over workers in the development and both test sets is $93.1\pm3.1$.
After pruning, it increases to $96.1\pm2.6$.
Table~\ref{tab:results:real_nlvr} shows human performance for each data split that has extra validations. 
Because  this process does not include the full dataset for each worker, it is not fully comparable to our evaluation results. 
However, it provides an estimate by balancing between averaging over many workers and having enough samples for each worker. 

%% file: evaluation_systems.tex
\section{Evaluation Systems}
\label{sec:eval_systems}

We evaluate several baselines and existing visual reasoning approaches using  \realnlvr. 
For all systems, we optimize for example-level accuracy.\footnote{System and learning details are available in Appendix~\ref{sec:app:systems}. } 

We measure the biases in the data using three baselines: (a) \system{Majority}: assign the most common label ($\truelabel$) to each example; (b) \system{Text}: encode the caption using a recurrent neural network~\cite[RNN; ][]{Elman:90rnn}, and use a multilayer perceptron to predict the truth value; and (c) \system{Image}: encode the pair of images using a convolutional neural network (CNN), and use a multilayer perceptron to predict the truth value.
The latter two estimate the potential of solving the task using only one of the two modalities.  

We use two baselines that consider both language and vision inputs. The \system{CNN+RNN} baseline concatenates the encoding of the text and images, computed similar to the \system{Text} and \system{Image} baselines, and applies a multilayer perceptron to predict a truth value. 
The \system{MaxEnt} baseline  computes features from the sentence and objects detected in the paired images. 
We detect the objects in the images using a Mask R-CNN model~\cite{He:17maskrcnn,Girshick:18detectron} pre-trained on the COCO detection task~\cite{Lin:14coco}.  We use a detection threshold of $0.5$. 
For each $n$-gram with a numerical phrase in the caption  and object class detected in the images, we compute features based on the number present in the $n$-gram and the detected object count. 
We create features for each image and for both together, and use these features in a maximum entropy classifier.

\begin{table*}[!h]
\begin{footnotesize}
\begin{center}
\begin{tabular}{|p{4.0cm}|c|c|c|c|} 
	\hline
	& \textbf{Train} & \textbf{Dev} & \textbf{Test-P} & \textbf{Test-U} \\ 
	\hline
	\system{Majority} (assign $\truelabel$) & $50.8 / 2.1$ & $50.9 / 3.9$ & $51.1 / 4.2$ & $51.4 / 4.6$ \\  \dline
    \system{Text} & \stdev{50.8}{0.0}$/$\stdev{2.1}{0.0} & \stdev{50.9}{0.0}$/$\stdev{3.9}{0.0} & $51.1/4.2$ & $51.4/4.6$ \\  \dline
	\system{Image}  & \stdev{60.1}{2.9}$/$\stdev{14.2}{4.2} & \stdev{51.6}{0.2}$/$\stdev{8.4}{0.8} & $51.9/7.4$ & $51.9/7.1$\\ \dline
	\system{CNN+RNN} & \stdev{94.3}{3.3}$/$\stdev{84.5}{10.2}& \stdev{53.4}{0.4}$/$\stdev{12.2}{0.7} & $52.4/11.0$ & $53.2/11.2$ \\ \dline
	\system{MaxEnt} & $89.4/73.4$ & $54.1/11.4$ & $\mathbf{54.8/11.5}$ & $\mathbf{53.5/12.0}$ \\ \hline \hline
\system{N2NMN}~\cite{Hu:17n2nmn}: & & & &\\
\system{N2NMN-Cloning} & \stdev{65.7}{25.8}$/$\stdev{30.8}{49.7} & \stdev{50.2}{1.0}$/$\stdev{5.7}{3.1} & -- & -- \\
\system{N2NMN-Tune} & \stdev{96.5}{1.6}$/$\stdev{94.9}{0.4} & \stdev{50.0}{0.7}$/$\stdev{9.8}{0.5} & --  & -- \\
\system{N2NMN-RL} & \stdev{50.8}{0.3}$/$\stdev{2.3}{0.3} & \stdev{51.0}{0.1}$/$\stdev{4.1}{0.3} & $51.1/5.0$ & $51.5 / 5.0$ \\ \dline
\film~\cite{Perez:17film} & \stdev{69.0}{16.9}$/$\stdev{32.4}{29.6} & \stdev{51.0}{0.4}$/$\stdev{10.3}{1.0} & $52.1/9.8$ & $53.0 / 10.6$ \\ \dline
\system{MAC}  &\multirow{2}{*}{\stdev{87.4}{0.8}$/$\stdev{64.0}{1.7}} & \multirow{2}{*}{\stdev{50.8}{0.6}$/$\stdev{11.0}{0.2}} & \multirow{2}{*}{$51.4/11.4$} & \multirow{2}{*}{$51.2/11.2$} \\
\cite{Hudson:18mac} & & & & \\
	\hline
	\hline
\system{Human} & -- & \stdev{96.2}{2.1}$/$-- & \stdev{96.3}{2.9}$/$-- & \stdev{96.1}{3.1}$/$-- \\ \hline
\end{tabular}
\end{center}
\end{footnotesize}
\caption{Performance (accuracy/consistency) on \realnlvr.}
\label{tab:results:real_nlvr}
\end{table*}

Several recent approaches to visual reasoning make use of modular networks (Section~\ref{sec:related}). 
Broadly speaking, these approaches predict a neural network layout from the input sentence by using a set of modules. The network is  used to reason about the image and text. 
The layout predictor may be trained: (a) using the formal programs used to generate synthetic sentences (e.g., in CLEVR), (b) using heuristically generated layouts from syntactic structures, or (c) jointly with the neural modules with latent layouts. 
Because sentences in \realnlvr  are human-written, no supervised formal programs are available at training time. 
We use two methods that do not require such formal programs: end-to-end neural module networks~\cite[N2NMN;][]{Hu:17n2nmn} and feature-wise linear modulation~\cite[\film;][]{Perez:17film}.
For N2NMN, we evaluate  three learning methods: (a) \system{N2NMN-Cloning}: using supervised learning with gold layouts; (b) \system{N2NMN-Tune}: using policy search after cloning; and (c) \system{N2NMN-RL}: using policy search from scratch. 
For \system{N2NMN-Cloning}, we construct layouts from constituency trees~\cite{Cirik:18}.
Finally, we evaluate the Memory, Attention, and Composition approach~\cite[\system{MAC};][]{Hudson:18mac}, which uses a sequence of attention-based steps. 
We modify \system{N2NMN}, \film, and \system{MAC} to process a pair of images by extracting image features from the concatenation of the pair. 

%% file: results.tex
\section{Experiments and Results}
\label{sec:experiments}

We use two metrics: accuracy and consistency. 
Accuracy measures the per-example prediction accuracy.
Consistency measures the proportion of unique sentences for which predictions are correct for all paired images~\cite{Goldman:17}.
For training and development results, we report  mean and standard deviation of accuracy and consistency over three trials as \stdev{\mu_{{\rm acc}}}{\sigma_{{\rm acc}}}$/$\stdev{\mu_{{\rm cons}}}{\sigma_{{\rm cons}}}.
The results on the test sets are generated by evaluating the model that achieved the highest accuracy on the development set.
For the N2NMN methods, we report test results only for the best of the three variants on the development set.\footnote{For reference, we also provide NLVR results in Table~\ref{tab:results:nlvr}, Appendix~\ref{sec:app:nlvr}.}

Table~\ref{tab:results:real_nlvr} shows results for \realnlvr.
\system{Majority} results demonstrate the data is fairly balanced. The results are slightly higher than perfect balance due to pruning (Sections~\ref{sec:validation} and~\ref{sec:split}). 
The \system{Text} and \system{Image} baselines perform similar to \system{Majority}, showing that both modalities are required to solve the task. 
\system{Text} shows identical performance to \system{Majority} because of how the data is balanced. 
The best performing system is the feature-based \system{MaxEnt} with the highest accuracy and consistency. 
\film performs best of the visual reasoning methods. 
Both \film and \system{MAC} show relatively high consistency. 
While almost all visual reasoning methods are able to fit the data, an indication of their high learning capacity, all generalize poorly. 
An exception is \system{N2NMN-RL}, which fails to fit the data, most likely due to the difficult task of policy learning from scratch. 
We also experimented with recent contextualized word embeddings to study the potential of stronger language models. 
We used a 12-layer uncased pre-trained BERT model~\cite{Devlin:19bert} with \film. 
We observed BERT provides no benefit, and therefore use the default embedding method for each model.

%% file: discussion.tex
\section{Conclusion}
\label{sec:conclusion}

We introduce the \realnlvr corpus for studying semantically-rich joint reasoning about photographs and natural language captions. 
Our focus on visually complex, natural photographs and human-written captions aims to  reflect the challenges of compositional visual reasoning better than existing corpora.
Our analysis shows that the language contains a wide range of linguistic phenomena including numerical expressions, quantifiers, coreference, and negation.
This demonstrates how our focus on complex visual stimuli and data collection procedure result in compositional and diverse language.
We experiment with baseline approaches and several methods for visual reasoning, which result in relatively low performance on \realnlvr. 
These results and our analysis exemplify the challenge that \realnlvr introduces to methods for visual reasoning.
We release training, development, and public test sets, and provide scripts to break down performance on the $800$ examples we manually analyzed (Section~\ref{sec:analysis}) according to the analysis categories. 
Procedures for evaluating on the unreleased test set and a leaderboard are available at  \href{http://lic.nlp.cornell.edu/nlvr/}{\tt http://lic.nlp.cornell.edu/nlvr/}.

%% file: ack.tex
\section*{Acknowledgments}
This research was supported by the NSF (CRII-1656998), a Google Faculty Award, a Facebook ParlAI Research Award, an AI2 Key Scientific Challenge Award, Amazon Cloud Credits Grant, and support from Women in Technology New York. 
This material is based on work supported by the National Science Foundation Graduate Research Fellowship under Grant No. DGE-1650441. 
We thank Mark Yatskar, Noah Snavely, and Valts Blukis for their comments and suggestions, the workers who participated in our data collection for their contributions, and the anonymous reviewers for their feedback.

%% file: supplementary.tex
\section{Frequently Asked Questions}
\label{sec:app:questions}

\paragraph{In what applications do you expect to see the kind of language \realnlvr allows to study?}

Composition of reasoning skills including counting, comparing, and reasoning about sets is critical for robotic agents following natural language instructions. 
Consider a robot on a factory floor or in a cluttered workshop following the instruction  \nlstring{get the two largest hammers from the toolbox at the end of the shelf}. 
Correctly following this instruction requires reasoning compositionally about object properties, comparisons between these properties, counts of objects, and spatial relations between observed objects. 
The language in \realnlvr reflects this type of linguistic reasoning. 
While the task we define does not use this kind of application directly, our data enables studying models that can understand this type of language. 

\paragraph{How can I use \realnlvr to build an end application?}
The task and data are not intended to directly develop an end application.
Our focus is on developing a task that drives research in vision and language understanding towards handling diverse set of reasoning skills.
It is critical to keep in mind that this dataset was not analyzed for social biases.
Researchers who wish to apply this work to an end product should take great care in considering what biases may exist.

\paragraph{Doesn't using a binary prediction task limit the ability to gain insight into model performance?}
Because our dataset contains both positive and negative image pairs for each sentence, we can measure consistency~\cite{Goldman:17}, which requires a model to predict each label correctly for each use of the sentence.
This metric requires generalization across at most four image pair contexts.

\paragraph{Why collect a new set of images rather than use existing ones like COCO~\cite{Lin:14coco}?}

Our goal was to achieve similar semantic diversity to NLVR, but using real images. 
Like NLVR, we use a sentence-writing task where sets of similar images are compared and contrasted. 
However, unlike NLVR, we do not have control over the image content, so cannot guarantee image sets where the content is similar enough (e.g., where the only difference is the direction in which the same animal is facing) such that the written sentence does not describe trivial image differences (e.g., the types of objects present).
In addition to image similarity within sets, we also prioritize image interestingness, for example images with many instances of an  object.
Existing corpora, including like COCO and ImageNet~\cite{Russakovsky:15}, were not constructed to prioritize interestingness as we define it, and are not comprised of sets of eight very similar images as required for our task.

\begin{enumerate}
    \item We select a set of $124$ ImageNet synsets which often appear in visually rich images.
    \item We generate search queries which result in visually rich images, e.g., containing multiple instances of a synset.
    \item We use a similar images tool to acquire sets of images with similar image content, for example containing the same objects in different relative orientations.
    \item We prune images which do not contain an example of the synset it was derived from.
    \item We apply a re-ranking and pruning procedure that prioritizes visually rich and interesting images, and prunes set which do not have enough interesting images.
\end{enumerate}

These steps result in a total of $17{,}685$ sets of eight similar, visually rich images.

\paragraph{Why use pairs of images instead of single images?}
We use pairs of images to elicit descriptions that reason over the pair of images in addition to the content within each image.
This setup supports, for example, comparing the two images, requiring that a condition holds in both images or in one but not the other, and performing set reasoning about the objects present in each image. 
This is analogous to the three-box setup in NLVR. 

\paragraph{Why allow workers to select the pairs themselves during sentence writing?}
We found that for some image pair selections, it was too difficult for workers to write a sentence which distinguishes the pairs. 
Allowing the workers to choose the pairs avoids this feasiblity issue. 

\paragraph{Why get multiple validations for development and test splits? }
This ensures the test  splits are of the highest quality and have minimal noise, as required for reliable measure of task performance. The additional annotatiosn also allow us to measure agreement and estimate human performance. 

\paragraph{How does the \realnlvr data compares to the NLVR data?}
NLVR and \realnlvr share the task of determining whether a sentence is true in a given visual context.
In NLVR, the visual input is synthetic and includes a handful of shapes and properties. 
In \realnlvr, each visual context is a pair of real photographs obtained from the web. 
Grounding sentences in image pairs rather than single images is related to NLVR's use of three boxes per image.

\paragraph{How does the \realnlvr data collection process compare to NLVR?}

We adapt the NLVR sentence-writing and validation tasks. 
However, rather than using four related synthetic images for writing, we use four pairs of real images. The pairing of images encourages set comparison. This was accomplished in NLVR through careful control of the generated image content, something that is not possible with real images. 
The NLVR image generation process is also controlled for the type of differences possible between images and the visual complexity, by ensuring the objects present in the selected and unselected images were the same.
This guarantees that the only differences are in the object configurations and distribution among the three boxes in each image.
Neither form of control is possible with real images. Instead, we rewrite the guidelines and develop a process to educate workers to follow them. 
In our process, we use the similar images tool to identify images that require linguistically-rich descriptions to distinguish. 
While using the similar images tool does not guarantee that the objects in the selected images are also present in the unselected images, our process successfully avoids this issue; in practice,  only around $13\%$ of examples take advantage of this by mentioning objects only present in the selected images. 

\paragraph{Can you summarize the key linguistic differences between  \realnlvr and NLVR?}

NLVR contains significantly\footnote{\label{fn:significance}Using a $\chi^2$ test with $p<0.05$.} more examples of hard cardinality, existential quantifiers, spatial relations, and prepositional attachment ambiguity.
\realnlvr contains significantly$^{\ref{fn:significance}}$ more examples of soft cardinality, universal quantifiers, coordination, coreference, and comparatives.
\realnlvr's descriptions are longer on average than NLVR ($14.8$ vs. $11.2$ tokens), and the vocabulary is much larger ($7{,}457$ vs. $262$ word types).
This demonstrates both the lexical diversity and challenges of understanding a wide range of image content in \realnlvr that are not present in NLVR. 
However, NLVR allows studying compositionality in isolation from lexical diversity, an intended feature of the dataset's design.
NLVR has also been used as a semantic parsing task, where images are represented as structured representations~\cite{Goldman:17}, a use case that is not possible with \realnlvr. 
NLVR remains a challenging dataset for visual reasoning; recent approaches have shown moderate improvements over the initial baseline performance, yet remain far from human accuracy, which we compute in Table~\ref{tab:results:nlvr}.

\paragraph{How does \realnlvr compare to existing visual reasoning datasets?}

Table~\ref{tab:dataset_comp} compares \realnlvr with several existing, related corpora. 
In the last several years there has been an increase in the number of datasets released for vision and language research.
One trend includes building datasets for compositional visual reasoning (SHAPES, CLEVR, CLEVR-Humans, ShapeWorld, NLVR, FigureQA, COG, and GQA), all of which use synthetic data either for at least one of the inputs. 
While \realnlvr requires related visual reasoning skills, it uses both real natural language and real visual inputs.

\paragraph{How does \realnlvr compare to recent attempts to avoid biases in vision and language datasets?}

Recently, several approaches were proposed to identify unintended biases present in vision-and-language tasks, such as the ability to answer a question without using the paired image~\cite{Zhang:16,Goyal:17,Li:17,Agrawal:17,Agrawal:18}.
The data collection process of \realnlvr is designed to automatically pair each sentence with both labels in different visual contexts. This makes \realnlvr robust to implicit linguistic biases. This is illustrated by our initial experiments with BERT, which have been shown to be extremely effective at capturing language patterns for various tasks~\cite{Devlin:19bert}. With our balanced data, using BERT does not help identifying and using language biases. 

\paragraph{Are the differences in the linguistic analysis between the datasets significant?}

We measure significance using a $\chi^2$ test with $p<0.05$. 
Our qualitative linguistic analysis shows several differences from VQA~\cite{Antol:15vqa} and GQA~\cite{Hudson:19gqa}.
\realnlvr contains significantly more examples of hard cardinality, soft cardinality, existential quantifiers, universal quantifiers, coordination, coreference, spatial relations, comparatives, negation, and preposition attachment ambiguity than both GQA and VQA.
However, VQA and GQA both contain significantly more examples of presupposition than \realnlvr.

\paragraph{Given your linguistic analysis, how does GQA compare to VQA?}

We found that the distribution of phenomena in VQA and GQA are roughly similar, with notable differences being significantly$^{\ref{fn:significance}}$ more examples of hard cardinality and coreference in VQA, and significantly$^{\ref{fn:significance}}$ more examples of universal quantifiers, coordination, and coordination and subordinating conjunction  attachment ambiguity in GQA.

\begin{table*}
\centering\footnotesize
\begin{tabular}{|m{3.6cm}|m{3.4cm}|m{3.2cm}|c|c|c|} \hline
\multicolumn{1}{|c|}{\multirow{2}{*}{\textbf{Dataset}}} &\multicolumn{1}{c|}{\multirow{2}{*}{\textbf{Task}}} & \multicolumn{1}{c|}{\textbf{Prevalent Linguistic}} & \textbf{Natural} & \textbf{Natural} \\ 
& & \multicolumn{1}{c|}{\textbf{Phenomena}} & \textbf{Language?} & \textbf{Images?} \\ \hline\hline
    \realnlvr & \multicolumn{1}{c|}{Binary Sentence Classification} &  (1) Hard and (2) soft cardinality; (3) existential and (4) universal quantifiers; (5) coordination; (6) coreference; (7) spatial relations; (8) presupposition; (9) preposition attachment ambiguity &  \multirow{2}{*}{\ding{52}} &  \multirow{2}{*}{\ding{52}} \\ \hline\hline
    VQA1.0~\cite{Antol:15vqa}, VQA-CP~\cite{Agrawal:17}, VQA2.0~\cite{Goyal:17} & \multicolumn{1}{c|}{Visual Question Answering}  &  (1) Hard cardinality; (2) existential quantifiers; (3) spatial relations; (4) presupposition & \ding{52}& \ding{52} \\ \dline
    NLVR~\cite{Suhr:17visual-reason} & \multicolumn{1}{c|}{Binary Sentence Classification} & (1) Hard and (2) soft cardinality; (3) existential quantifiers; (4) coordination; (5) spatial relations; (6) presupposition; (7) preposition attachment ambiguity &  \multirow{2}{*}{\ding{52}} & \\ \dline
    GQA~\cite{Hudson:19gqa} & \multicolumn{1}{c|}{Visual Question Answering} & (1) Existential quantifiers; (2) coordination; (3) spatial relations; (4) presupposition & & \ding{52} \\ \hline
    \multicolumn{5}{c}{} \\
\end{tabular}

\begin{tabular}{|m{5.5cm}|m{3.4cm}|c|c|} \hline
\multicolumn{1}{|c|}{\multirow{2}{*}{\textbf{Dataset}}} &\multicolumn{1}{c|}{\multirow{2}{*}{\textbf{Task}}} & \textbf{Natural} & \textbf{Natural} \\ 
& & \textbf{Language?} & \textbf{Images?} \\ \hline    
    SAIL~\cite{MacMahon:06} & \multicolumn{1}{c|}{Instruction Following}  & \ding{52} & \\ \dline
    \citet{Mitchell:10} & \multicolumn{1}{c|}{Referring Expression Resolution} & \ding{52} & \\ \dline
    \citet{Matuszek:12} & \multicolumn{1}{c|}{Referring Expression Resolution} & \ding{52} & \\ \dline
    \citet{FitzGerald:13} & \multicolumn{1}{c|}{Referring Expression Generation} & \ding{52} & \\ \dline
    VQA (Abstract)~\cite{Zitnick:13abstract} & \multicolumn{1}{c|}{Visual Question Answering} & \ding{52} & \\ \dline
    ReferItGame~\cite{Kazemzadeh:14} &\multicolumn{1}{c|}{Referring Expression Resolution} & \ding{52}& \ding{52} \\ \dline
     SHAPES~\cite{Andreas:16nmn} & \multicolumn{1}{c|}{Visual Question Answering} & & \\ \dline
   \citet{Bisk:16dataset} &  \multicolumn{1}{c|}{Instruction Following} & \ding{52} & \\ \dline
\textsc{MSCOCO}~\cite{Chen:16coco} & \multicolumn{1}{c|}{Caption Generation}  & \ding{52}& \ding{52}\\ \dline
    Google RefExp~\cite{Mao:16} &  \multicolumn{1}{c|}{Referring Expression Resolution} & \ding{52}& \ding{52} \\ \dline 
     \textsc{Room-to-Room}~\cite{Anderson:17} & \multicolumn{1}{c|}{Instruction Following} & \ding{52}  & \ding{52} \\ \dline
  Visual Dialog~\cite{Das:17visdial} & \multicolumn{1}{c|}{Dialogue Visual Question Answering} & \ding{52}& \ding{52} \\ \dline
    CLEVR~\cite{Johnson:16clevr} & \multicolumn{1}{c|}{Visual Question Answering} & & \\ \dline
     CLEVR-Humans~\cite{Johnson:17iep} & \multicolumn{1}{c|}{Visual Question Answering} & \ding{52} & \\ \dline
    TDIUC~\cite{Kafle:17tdiuc} & \multicolumn{1}{c|}{Visual Question Answering}  & \ding{52}& \ding{52} \\ \dline
    ShapeWorld~\cite{Kuhnle:17} & \multicolumn{1}{c|}{Binary Sentence Classification} & & \\ \dline
   FigureQA~\cite{Kahou:17figureqa} & \multicolumn{1}{c|}{Visual Question Answering} & & \\ \dline
    TVQA~\cite{Lei:18tvqa} & \multicolumn{1}{c|}{Video Question Answering} & \ding{52}& \ding{52} \\ \dline
    \textsc{Lani} \& \textsc{CHAI}~\cite{Misra:18goalprediction} & \multicolumn{1}{c|}{Instruction Following} & \ding{52} & \ding{52}  \\ \dline
    Talk the Walk~\cite{DeVries:18ttw} & \multicolumn{1}{c|}{Dialogue Instruction Following} & \ding{52}& \ding{52}\\ \dline
    \multirow{2}{*}{COG~\cite{Yang:18cog}} & \multicolumn{1}{c|}{Visual Question Answering;} & & \\ 
    & \multicolumn{1}{c|}{Instruction Following} & & \\ \dline
  VCR~\cite{Zellers:18vcr} &\multicolumn{1}{c|}{Visual Question Answering}  & \ding{52} & \ding{52}  \\ \dline
    TallyQA~\cite{Acharya:18} &\multicolumn{1}{c|}{Visual Question Answering}  & \ding{52} & \ding{52}  \\ \dline
    \multirow{2}{*}{\textsc{Touchdown}~\cite{Chen:19touchdown}} & \multicolumn{1}{c|}{Instruction Following;} & \multirow{2}{*}{\ding{52}} & \multirow{2}{*}{\ding{52}}  \\
    & \multicolumn{1}{c|}{Spatial Description Resolution} & & \\ \dline
    \textsc{COCO-BISON}~\cite{Hu:19bison} & \multicolumn{1}{c|}{Binary Image Selection}&\ding{52} &\ding{52} \\ \dline
    SNLI-VE~\cite{Xie:19} & \multicolumn{1}{c|}{Visual Entailment} & \ding{52}& \ding{52} \\ \hline
\end{tabular}
\caption{Comparison between \realnlvr and  existing datasets for language and vision  research. The top table details prevalent linguistic phenomena in some of the most related datasets according to our analysis, listing each  linguistic phenomenon with at least $10\%$ representation as prevalent. For each dataset, we count the number of prevalent phenomena. \realnlvr has the broadest representation. The bottom table lists other tasks in language and vision.}
\label{tab:dataset_comp}
\end{table*}

\section{Data Collection Details}
\label{sec:app:data_collection}
\paragraph{Image Collection}

We consider the images of each search query in the order of the search results. 
For each result associated with a set of similar images, we save the URL of the result image and the URLs of the fifteen most similar images, giving us a set of sixteen images.
We skip and ignore URLs from a hand-crafted list of stock photo domains; images from these domains include large, distracting watermarks.
We stop after observing $60$ result images, saving $30$ sets of image URLs, or observing five consecutive results that do not have similar images.\footnote{For collective nouns and the numerical phrase \texttt{two <synset>}, we instead observe at most $100$ top images or save at most $60$ sets.}

After downloading a set of $16$ URLs of related images (Section~\ref{sec:image_collection}), we automatically prune the images.
We remove any broken URLs or any URLS that appeared in other previously-downloaded sets from the same search query. 
We remove downloaded images smaller than $200 \times 200$ pixels.
We apply basic duplicate removal by removing any images which are exact duplicates of a previously-downloaded image in the set. 
This automatic pruning may result in image sets consisting of fewer than $16$ images.
We discard any sets after this stage with fewer than $8$ images.

\begin{table*}[!h]
\centering\footnotesize
\begin{tabular}{|p{5.7cm}|p{9.4cm}|} \hline
\textbf{What to avoid} & \textbf{Example of erroneous sentence} \\ \hline
Subjective opinions & \nlstring{The dog's fur has a nice color pattern.} \\ \dline
Discussing properties of the photograph & \nlstring{In both images, the cat's paw is cropped out of the photo.} \\ \dline
Mentioning text in the photograph & \nlstring{Both trains are numbered 72.} \\ \dline
Mentioned object not present in unselected pairs & \nlstring{There is a cup on top of a chair.} -- for a set of images where the selected pairs contain a chair, but the unselected pairs do not. \\ \hline \hline
Mentioning the presence of a single object & \nlstring{There is a hammer.} \\ \dline
Disjunction on images in the pair & \nlstring{The left image contains a penguin, and the right image contains a rock.} \\ \hline
\end{tabular}
\caption{Types of sentences workers are discouraged from writing. The bottom two  are permissible as long as the sentence includes other kinds of reasoning.}
\label{tab:sentence_guidelines}
\end{table*}

\paragraph{Sentence Writing} 

Table~\ref{tab:sentence_guidelines} shows the types of sentences we ask workers to avoid in their writing.
Analysis of $100$ sentences from the development set shows that almost all sentences follow our guidelines, only $13\%$ violate our guidelines. 
The most common violation was mentioning an object not present in the unselected images.
Such sentences can trivially be labeled as $\falselabel$ in the context of the unselected pairs, as the mentioned object will not be present. 
In the context of the selected pairs, however, a model must still perform compositional joint reasoning about the sentence and the image pair to determine whether the label should be $\truelabel$ at test time. This is because the sentence often includes additional constraints. 
The bottom example in Table~\ref{tab:extra_examples} illustrates this violation.
A system may easily determine that because neither a hole nor a golf flagpole are present in either image, the sentence is $\falselabel$.
However, if these objects were present, the system must reason about counts and spatial relations of the mentioned objects to verify that the sentence is $\truelabel$.

\paragraph{Data Collection Management}
We use two qualification tasks. 
For the set construction and sentence writing tasks, we qualify workers by first showing six tutorial questions about the guidelines and task.
We then ask them to validate guidelines for nineteen sentences across two sets of four pre-selected image pairs, and to complete a single sentence-writing task for  pre-selected image pairs. 
We validate the written sentence by hand. 
We qualify  workers for validation with eight pre-selected validation tasks. 

We use a bonus system to encourage workers to write linguistically diverse sentences. 
We conduct sentence writing in rounds. After each round, we sample twenty sentences
 for each worker from that round. 
If at least $75\%$ of these sentences follow the guidelines, they receive a bonus for each sentence written during the last round.
If between $50\%$ and $75\%$ follow our guidelines, they receive a slightly lower bonus. 
This  encourages workers to follow the guidelines more closely. 
In addition, each worker initially only has access to a limited pool of sentence-writing tasks. Once  they successfully complete an evaluation round where at least $75\%$ of their sentences followed the guidelines, they get access to the entire pool of tasks.

\begin{table}[t]
\begin{footnotesize}
\begin{center}
\begin{tabular}{|l|c|c|} \hline
& \textbf{Cost} & \textbf{Unique Workers} \\ \hline
Image Pruning & \$1,310.76 & $53$\\ \dline
Set Construction & \$1,798.84 & $46$ \\ \dline
Sentence Writing & \$9,570.46 & $99$ \\ \dline
Validation & \$6,452.93 & $125$ \\ \hline
Total & \$19,132.99 & $167$ \\ \hline
\end{tabular}
\end{center}
\end{footnotesize}
\caption{Cost and worker statistics.}
\label{tab:costs}
\end{table}

Table~\ref{tab:costs} shows the costs and number of  workers per task.
The final cost per unique sentence in our dataset is \$0.65; the cost per example is \$0.18.

\section{Additional Data Analysis}
\label{sec:app:analysis}

\begin{table*}[!h]
\begin{footnotesize}
\begin{center}
\begin{tabular}{|l|c|l|}\hline
\textbf{Required Reasoning} & \textbf{\%} & \textbf{Example from \realnlvr} \\ \hline
Exactly one image & $3$ & \nlstring{\textbf{Only one image} shows warthogs butting heads.} \\ \dline
Existential quantification & $19$ &  \nlstring{\textbf{In one image}, hyenas fight with a big cat.} \\ \dline
Universal quantification & $11$ &  \nlstring{There are people walking \textbf{in both images}.} \\ \dline
Explicit reference to left and/or right image & $26.5$ & \nlstring{\textbf{The left image} contains exactly two dogs.} \\ \dline
Comparison between images & $6$ & \nlstring{There are \textbf{more} mammals in \textbf{the image on the right}.} \\ \hline
\end{tabular}
\end{center}
\end{footnotesize}
\caption{Types of reasoning over the pair of images required in \realnlvr, including the proportion of examples requiring each type and an example.}
\label{tab:image_refs}
\end{table*}

\paragraph{Synsets}
\input{synset_dist}
Figure~\ref{fig:synset_dist} shows the counts of examples per synset in the training and development sets.

\paragraph{Image Pair Reasoning} We use a $200$-sentence subset of the sentences analyzed in Table~\ref{tab:cats} to analyze what types of  reasoning are required over the two images (Table~\ref{tab:image_refs}). 
We observe that sentences commonly use the pair structure used to display the images: 
$11\%$ of sentences require that a property to hold in both images, 
$19\%$  simply require that a property holds in at least one image, and 
$26.5\%$ of sentences require a property to be true in the left or right images specifically.
The pair is also used for comparison, with $6\%$ of sentences requiring comparing properties of the two images. 
Finally, $39.5\%$ of sentences simply state a property that must be true across the image pair, e.g., \nlstring{One sliding door is closed}.

\section{Results on NLVR}
\label{sec:app:nlvr}
\begin{table*}[!h]
\begin{footnotesize}
\begin{center}
\begin{tabular}{|p{4.7cm}|c|c|c|c|} 
	\hline
	& \textbf{Train} & \textbf{Dev} & \textbf{Test-P} & \textbf{Test-U} \\ 
	\hline
	\system{Majority} (assign $\truelabel$) & $56.4 / $-- & $55.3 / $--  & $56.2 / $--  & $55.4 / $--  \\  \dline
    \system{Text} & \stdev{58.4}{0.6}$/$-- & \stdev{56.6}{0.5}$/$-- & \stdev{57.2}{0.6}$/$-- & \stdev{56.2}{0.4}$/$--  \\ \dline
	\system{Image}  & \stdev{56.8}{1.3}$/$-- & \stdev{55.4}{0.1}$/$-- & \stdev{56.1}{0.3}$/$--  & \stdev{55.3}{0.3}$/$--  \\ \dline
	\system{CNN+RNN} & \stdev{58.9}{0.2}$/$--   & \stdev{56.6}{0.3}$/$--   & \stdev{58.0}{0.3}$/$--  & \stdev{56.3}{0.6} $/$--  \\  \dline
\system{NMN} & \stdev{98.4}{0.6}$/$-- & \stdev{63.1}{0.1}$/$-- & \stdev{66.1}{0.4}$/$-- & \stdev{62.0}{0.8}$/$-- \\ \hline \hline%
\system{CNN-BiATT} & \multirow{2}{*}{--} & \multirow{2}{*}{$66.9/$--} & \multirow{2}{*}{$69.7/$--} & \multirow{2}{*}{$\mathbf{66.1 / 28.9}$} \\ 
 \cite{Tan:18cnn-biatt} & & & & \\ \dline
\system{W-MemNN}~\cite{Pavez:18} & -- & $65.6/$--  & $65.8/$-- &  -- \\  \dline
\system{CMM}~\cite{Yao:18cmm} & -- & $68.0/$-- & $\mathbf{69.9}/$-- & -- \\ \hline \hline
\system{N2NMN}~\cite{Hu:17n2nmn}: & & & & \\
\system{N2NMN-Cloning} & \stdev{95.6}{1.3}$/$\stdev{79.9}{4.7} & \stdev{57.9}{1.1}$/$\stdev{9.7}{0.8} & -- & -- \\
\system{N2NMN-Tuning} & \stdev{97.5}{0.4}$/$\stdev{92.7}{2.6} & \stdev{58.7}{1.4}$/$\stdev{11.6}{0.8} & --  & --\\ 
\system{N2NMN-RL} & \stdev{95.4}{2.4}$/$\stdev{81.2}{10.6} & \stdev{65.3}{0.4}$/$\stdev{16.2}{1.5} & $69.1/20.7$ & $66.0 / 17.7$ \\ \dline
\film~\cite{Perez:17film} & \stdev{95.5}{0.4}$/$\stdev{84.6}{2.7} & \stdev{60.1}{1.2}$/$\stdev{14.6}{1.3} & $62.2/18.4$ & $61.2/18.1$\\
	\dline
\system{MAC} & \multirow{2}{*}{\stdev{64.2}{4.7}$/$\stdev{12.6}{0.2}} & \multirow{2}{*}{\stdev{55.4}{0.5}$/$\stdev{7.4}{0.6}} & \multirow{2}{*}{$57.6/11.7$} & \multirow{2}{*}{$54.3/8.6$} \\ 
\cite{Hudson:18mac}  & & & & \\ \hline \hline
\system{Human} (approximation)& -- & \stdev{94.6}{3.5}$/$-- & \stdev{95.4}{3.4}$/$-- & \stdev{94.9}{3.6}$/$-- \\ \hline
\end{tabular}
\end{center}
\end{footnotesize}
\caption{Performance (accuracy/consistency) on NLVR.}
\label{tab:results:nlvr}
\end{table*}

Table~\ref{tab:results:nlvr} shows previously published results using raw images in NLVR from \citet{Suhr:17visual-reason} and more recent approaches.\footnote{Not all previously evaluated methods report consistency.}
We also report results for visual reasoning systems originally developed for CLEVR. 
We compute human performance for each split of the data using the procedure described in Section~\ref{sec:human_performance}; a threshold of $100$ covers $100\%$ of annotators.
\system{NMN}~\cite{Andreas:16nmn}, \system{N2NMN}, and \film achieve the best results for methods that were not developed using NLVR. 
However, both perform worse than \system{CNN-BiATT}~\cite{Tan:18cnn-biatt} and CMM~\cite{Yao:18cmm}, which were developed originally using NLVR.\footnote{Consistency for \system{CNN-BiATT} was taken from the NLVR leaderboard.}

\section{Implementation Details}
\label{sec:app:systems}
For the \textsc{Text}, \textsc{Image}, and \text{CNN+RNN} baselines, we first compute a representation of the input(s). We then process this representation using a multilayer perceptron (MLP).
The MLP's output is used to predict a distribution over the two labels using a $\mathrm{softmax}$.
The MLP includes learned bias terms and $\mathrm{ReLu}$ nonlinearities on the output of each layer, except the last one. 
In all cases, the layer sizes of the MLP follow the series $[8192, 4096, 2048, 1024, 512, 256, 128, 64, $ $32, 16, 2]$.

\subsection{Single Modality}
\label{sec:single_modality_sup}
\paragraph{\system{Text}}
The caption's representation is computed  using an RNN encoder.
We use $300$-dimensional GloVe vectors trained on Common Crawl as word embeddings~\cite{Pennington:14glove}.
We encode the caption using a single-layer long short-term memory~\cite[LSTM,][]{Hochreiter:97lstm} RNN  of size $4096$.
The hidden states of the caption are averaged and processed with the MLP described above to predict the  truth value.

\paragraph{\system{Image}}
The image pair's representation is computed by extracting features from a pre-trained model.
We resize and pad each image with whitespace to a size of $530 \times 416$ pixels, which is the size of the image displayed to the workers during sentence-writing.
Each padded image is resized to $224 \times 224$ and passed through a ResNet-152 pre-trained model~\cite{He:16resnet}.
The features from the final layer before classification are extracted for each image and concatenated.
This representation is processed with the MLP described above to predict a truth value.

\subsection{Image and Text Baselines}
\paragraph{\system{CNN+RNN}}
The caption and image pair are encoded as described in Appendix~\ref{sec:single_modality_sup}, then concatenated and passed through the MLP described above to predict a truth value.

\paragraph{\system{MaxEnt}}
We use $n$-grams where \mbox{$2 \leq n \leq 6$}.
We train a maximum entropy classifier with Megam.\footnote{\href{https://www.umiacs.umd.edu/~hal/megam}{\tt https://www.umiacs.umd.edu/\textasciitilde hal/megam}}

\subsection{Module Networks}
\paragraph{End-to-End Neural Module Networks}

We use the publicly available implementation.\footnote{\href{https://github.com/ronghanghu/n2nmn}{\tt https://github.com/ronghanghu/n2nmn}}
The model parameters used for \realnlvr are the same as those used for the original experiments on VQA.
We use GloVe vectors of size $300$ to embed words~\cite{Pennington:14glove}.
The model parameters used for NLVR are the same as those used for the original \textsc{N2NMN} experiments on CLEVR.
This includes learning word embeddings from scratch and embedding images using the \textit{pool5} layer of VGG-16 trained on ImageNet~\cite{Simonyan:15,Hu:17n2nmn}.
The two paired images are resized and padded with white space to size $530 \times 416$, then concatenated horizontally and resized to a single image of $448 \times 448$ pixels. 
The resulting image is embedded using the \textit{res5c} layer of ResNet-152 trained on ImageNet~\cite{He:16resnet,Hu:17n2nmn}.

\paragraph{\film} 
We use the publicly available implementation.\footnote{\href{https://github.com/ethanjperez/film}{\tt https://github.com/ethanjperez/film}} 
For \realnlvr, we first resize and pad both images with whitespace to images of size $530 \times 416$.
The two images are concatenated horizontally and resized to a single image of $224 \times 224$ pixels.
This image is passed through a ResNet-101 pretrained model and the features from the \textit{conv4} layer are extracted~\cite{He:16resnet,Perez:17film}.
For NLVR, we resize images to $224 \times 224$ and use the raw pixels directly.
The parameters of the models are the same as described in Perez et al.~\shortcite{Perez:17film}'s experiments on featurized images, except for the following: RNN hidden size of $1096$, classifier projection dimension of size $256$, final MLP hidden size of $512$, and $28$ feature maps. 
Using the original parameters did not result in significant differences in accuracy, while updates using our parameters were computed faster and the computation graph used less memory.

\subsection{\textsc{MAC}}
We use the implementation provided online.\footnote{\href{https://github.com/stanfordnlp/mac-network}{\tt https://github.com/stanfordnlp/\hspace{0pt}mac-network}}
For experiments on \realnlvr, we adapt the image processing procedure.
Both images are resized and padded with white space to images of size $530 \times 416$, then concatenated horizontally and resized to $224 \times 224$ pixels. 
We use the same image featurization approach used in \citet{Hudson:18mac}.
For experiments on NLVR, we use the NLVR configuration provided in the repository.

\subsection{Training}
For the \system{Text}, \system{Image}, and \system{CNN+RNN} methods on \realnlvr, we perform updates using \textsc{Adam}~\cite{Kingma:14adam} with a global learning rate of $0.0001$.
The weights and biases are initialized by sampling uniformly from $[-0.1, 0.1]$.
All fully-connected and output layers use a learned bias term.
For \system{MAC}, we use the same training setup as described in \citet{Hudson:18mac}, stopping early based on performance over the development set.
For all other experiments, we use early stopping with patience, where patience is initially set to a constant and multiplied $1.01$ at each epoch the validation accuracy improves over a global maximum. 
We use $5\%$ of the training data as a validation set, which is not used to update model parameters.
We choose a validation set such that unique sentences do not appear in both the validation and training sets.
For \film  and \system{N2NMN}, we set the initial patience to $30$.
For \system{Text}, \system{Image} and \system{CNN+RNN} baselines, initial patience was set to $10$.
For \system{MaxEnt}, we use at most $100$ epochs.

\section{Additional Examples}
\input{extra_examples}
Table~\ref{tab:extra_examples} includes additional examples sampled from the training and development sets of \realnlvr, as well as license information for each image.
All images in this paper were sampled from websites known for hosting non-copyrighted images, for example Wikimedia.

\section{Lisence Information}
\label{sec:app:license}
Tables~\ref{tab:sup_license},~\ref{tab:sup_license2},~\ref{tab:sup_license3}, and~\ref{tab:sup_license3} detail license and attribution information for the images included in the main paper.

\newcommand{\tabimage}[1]{\centering\vspace{2pt}\includegraphics[width=0.8\linewidth]{{#1}}\vspace{-1pt}}

\begin{table}
\footnotesize
\centering
\begin{tabular}{|m{2.8cm}|p{3.3cm}|} \hline
\textbf{Image} & \textbf{Attribution and License} \\ \hline
\tabimage{figs/ex0_0.jpg} & MemoryCatcher \newline (CC0) \\ \dline
\tabimage{figs/ex0_1.jpg} & Calabash13\newline  (CC BY-SA 3.0)\\ \dline
\tabimage{figs/acorns_1.jpg} & Charles Rondeau\newline  (CC0) \\ \dline
\tabimage{figs/acorns_6.jpg} & Andale \newline (CC0) \\ \hline
\end{tabular}
\caption{License information for the images in Figure~\ref{fig:examples}.}
\label{tab:sup_license}
\end{table}

\begin{table}
\footnotesize
\centering
\begin{tabular}{|m{2.8cm}|p{3.3cm}|} \hline
\textbf{Image} & \textbf{Attribution and License} \\ \hline
\tabimage{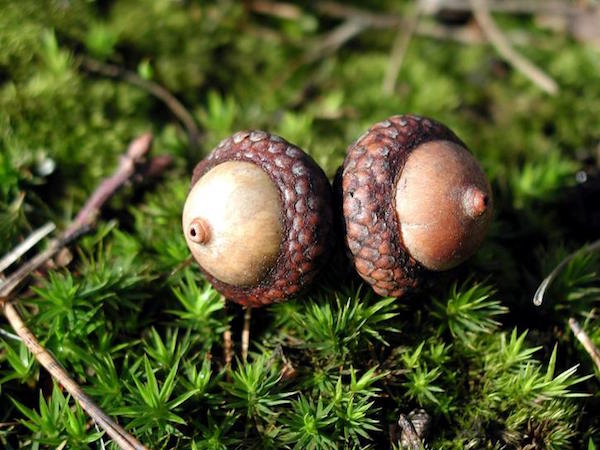} & Hagerty Ryan, USFWS  \newline (CC0) \\ \dline
\tabimage{figs/acorns_1.jpg} & Charles Rondeau  \newline (CC0) \\ \dline
\tabimage{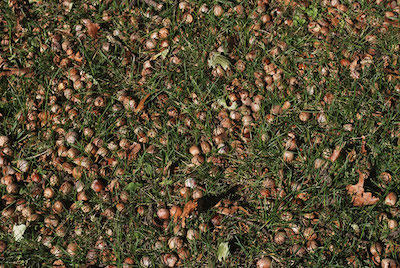} & Peter Griffin  \newline (CC0) \\ \dline
\tabimage{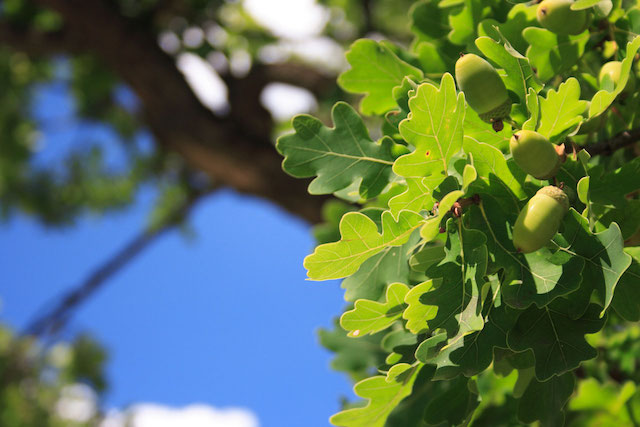} & Petr Kratochvil  \newline (CC0) \\ \dline
\tabimage{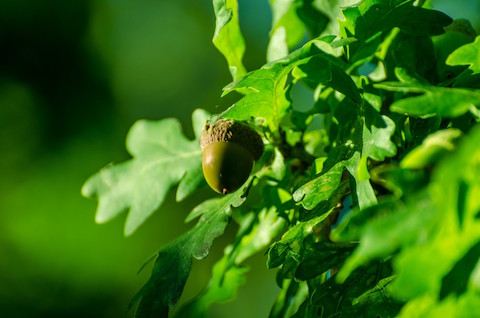} & George Hodan  \newline (CC0) \\ \dline
\tabimage{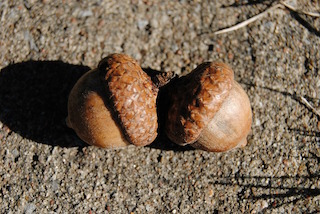} & Charles Rondeau  \newline (CC0) \\ \dline
\tabimage{figs/acorns_6.jpg} & Andale \newline  (CC0) \\ \dline
\tabimage{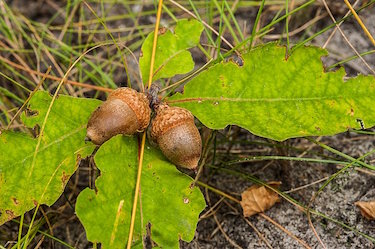} & Maksym Pyrizhok \newline  (PDP) \\ \dline
\tabimage{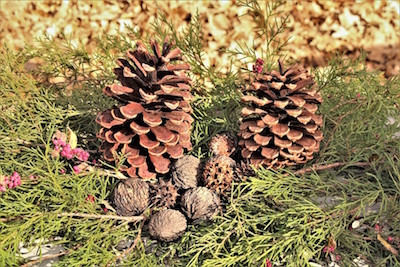} & Sheila Brown \newline  (CC0) \\ \dline
\tabimage{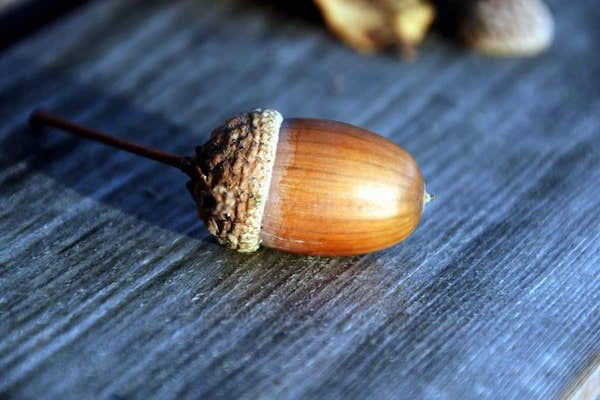} & ulleo \newline  (CC0) \\ \hline
\end{tabular}
\caption{License information for the images in Figure~\ref{fig:collection}.}
\label{tab:sup_license2}
\end{table}

\begin{table}
\footnotesize
\centering
\begin{tabular}{|m{2.8cm}|p{3.3cm}|} \hline
\textbf{Image} & \textbf{Attribution and License} \\ \hline
\tabimage{figs/vulture_0.JPG} & JerryFriedman \newline (CC0) \\ \dline
\tabimage{figs/vulture_1.jpg} & Eric Kilby \newline (CC BY-SA 2.0)\\ \dline
\tabimage{figs/vulture_2.jpg} & Angie Garrett \newline (CC BY 2.0) \\ \dline
\tabimage{figs/vulture_3.jpg} & Ben HaTeva \newline (CC BY-SA 2.5)\\ \dline
\tabimage{figs/train0.jpg} & Manfred Kopka \newline (CC BY-SA 4.0)\\ \dline
\tabimage{figs/train1.jpg} & Aubrey Dale \newline (CC BY-SA 2.0) \\ \dline
\tabimage{figs/train2.jpg} & Albert Bridge \newline (CC BY-SA 2.0) \\ \dline
\tabimage{figs/train3.jpg} & Randwick \newline (CC BY-SA 3.0)\\ \dline
\tabimage{figs/goose0.jpg} & Alexas\_Fotos \newline (Pixabay) \\ \dline
\tabimage{figs/goose1.jpg} & Alexas\_Fotos \newline (Pixabay) \\ \dline
\tabimage{figs/goose2.jpg} & Ralph Daily \newline (CC BY 2.0) \\ \dline
\tabimage{figs/goose3.jpg} & hobbyknipse \newline (Pixabay) \\ \hline
\end{tabular}
\caption{License information for the images in Table~\ref{tab:paired_examples}.}
\label{tab:sup_license3}
\end{table}

\begin{table}\centering\footnotesize
\begin{tabular}{|m{2.8cm}|p{4cm}|} \hline
\textbf{Image} & \textbf{Attribution and License} \\ \hline
\tabimage{figs/ex_many.jpg} & Nedih Limani \newline (CC BY-SA 3.0) \\ \dline
\tabimage{figs/ex_interact.jpg} & Jean-Pol GRANDMONT \newline (CC BY-SA 3.0) \newline  \\ \dline
\tabimage{figs/ex_activity.jpg} & Scott Robinson \newline (CC BY 2.0) \\ \dline
\tabimage{figs/ex_diverse.jpg} & Tokumeigakarinoaoshima  \newline (CC0 1.0) \\ \dline
\tabimage{figs/ex_neg_1.jpg} & CSIRO \newline (CC BY 3.0) \\ \dline
\tabimage{figs/ex_neg_2.jpg} & Dan90266  \newline (CC BY-SA 2.0) \\ \dline
\tabimage{figs/ex_neg_7.jpg} & Raimond Spekking \newline (CC BY-SA 4.0) \\ \dline
\tabimage{figs/ex_neg_8.jpg} & SamHolt6  \newline (CC BY-SA 4.0) \\ \hline
\end{tabular}
\caption{License information for the images in Table~\ref{tab:interesting_criteria}.}
\label{tab:sup_license4}
\end{table}

%% file: synset_dist.tex
\begin{figure}[t]
\begin{footnotesize}
\begin{center}
\begin{tikzpicture}
\begin{axis}[ybar stacked,
    width=0.95\columnwidth,
        height=0.51\columnwidth,
        bar width=0.05em,
        ytick={0,250,500,750,1000,1250},
        xmin=-1,
        xmax=125,
        ymin=0,
        ymax=1250,
        xticklabels=\empty,
        legend pos=north west]
\addplot coordinates{
(0,189)
(1,187)
(2,211)
(3,227)
(4,289)
(5,362)
(6,375)
(7,398)
(8,431)
(9,437)
(10,455)
(11,478)
(12,489)
(13,495)
(14,496)
(15,500)
(16,517)
(17,516)
(18,526)
(19,526)
(20,544)
(21,545)
(22,547)
(23,549)
(24,584)
(25,562)
(26,562)
(27,562)
(28,569)
(29,569)
(30,565)
(31,568)
(32,570)
(33,568)
(34,572)
(35,574)
(36,577)
(37,584)
(38,600)
(39,591)
(40,600)
(41,600)
(42,615)
(43,618)
(44,616)
(45,624)
(46,636)
(47,623)
(48,644)
(49,649)
(50,654)
(51,654)
(52,665)
(53,662)
(54,661)
(55,664)
(56,671)
(57,673)
(58,675)
(59,679)
(60,684)
(61,689)
(62,697)
(63,698)
(64,701)
(65,710)
(66,706)
(67,707)
(68,718)
(69,712)
(70,722)
(71,718)
(72,725)
(73,726)
(74,724)
(75,725)
(76,729)
(77,732)
(78,743)
(79,753)
(80,746)
(81,745)
(82,752)
(83,755)
(84,771)
(85,778)
(86,778)
(87,794)
(88,787)
(89,794)
(90,806)
(91,830)
(92,817)
(93,816)
(94,817)
(95,821)
(96,832)
(97,848)
(98,852)
(99,861)
(100,859)
(101,869)
(102,891)
(103,906)
(104,908)
(105,913)
(106,908)
(107,917)
(108,923)
(109,926)
(110,944)
(111,956)
(112,962)
(113,970)
(114,988)
(115,997)
(116,1049)
(117,1087)
(118,1093)
(119,1096)
(120,1095)
(121,1110)
(122,1166)
(123,1176)
};
\addlegendentry{Train}
\addplot coordinates{
(0,14)
(1,18)
(2,12)
(3,22)
(4,27)
(5,34)
(6,38)
(7,38)
(8,38)
(9,47)
(10,48)
(11,43)
(12,45)
(13,53)
(14,54)
(15,50)
(16,39)
(17,52)
(18,46)
(19,56)
(20,52)
(21,55)
(22,53)
(23,55)
(24,31)
(25,56)
(26,57)
(27,61)
(28,56)
(29,58)
(30,62)
(31,59)
(32,57)
(33,60)
(34,59)
(35,58)
(36,57)
(37,55)
(38,53)
(39,62)
(40,64)
(41,64)
(42,54)
(43,53)
(44,59)
(45,58)
(46,46)
(47,62)
(48,58)
(49,64)
(50,62)
(51,62)
(52,57)
(53,60)
(54,62)
(55,66)
(56,61)
(57,61)
(58,61)
(59,62)
(60,63)
(61,58)
(62,58)
(63,62)
(64,66)
(65,57)
(66,62)
(67,61)
(68,54)
(69,63)
(70,57)
(71,62)
(72,61)
(73,61)
(74,63)
(75,64)
(76,61)
(77,63)
(78,61)
(79,51)
(80,59)
(81,62)
(82,59)
(83,61)
(84,55)
(85,64)
(86,66)
(87,57)
(88,66)
(89,59)
(90,66)
(91,46)
(92,59)
(93,63)
(94,65)
(95,62)
(96,65)
(97,61)
(98,58)
(99,60)
(100,62)
(101,61)
(102,64)
(103,54)
(104,57)
(105,59)
(106,67)
(107,64)
(108,65)
(109,65)
(110,62)
(111,55)
(112,57)
(113,57)
(114,60)
(115,64)
(116,60)
(117,62)
(118,62)
(119,61)
(120,64)
(121,63)
(122,65)
(123,65)
};
\addlegendentry{Dev}
\end{axis}
\end{tikzpicture}
\end{center}
\end{footnotesize}
\caption{Number of examples per synset, sorted by number of examples in each synset.}
\label{fig:synset_dist}
\end{figure}
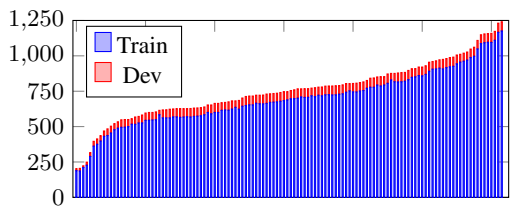

%% file: extra_examples.tex
\newcommand{\imgpair}[2]{\centering\vspace{0.4em}\frame{\colorbox{gray!20}{\frame{\includegraphics[height=11.5ex]{{#1}}}~{\frame{\includegraphics[height=11.5ex]{{#2}}}}}}}

\begin{table*}
    \centering\footnotesize
    \begin{tabular}{|m{7cm}|m{5cm}|c|} \hline
    \multicolumn{1}{|c|}{\textbf{Image Pair}} & \multicolumn{1}{c|}{\textbf{Sentence}} & \textbf{Label} \\ \hline
       \imgpair{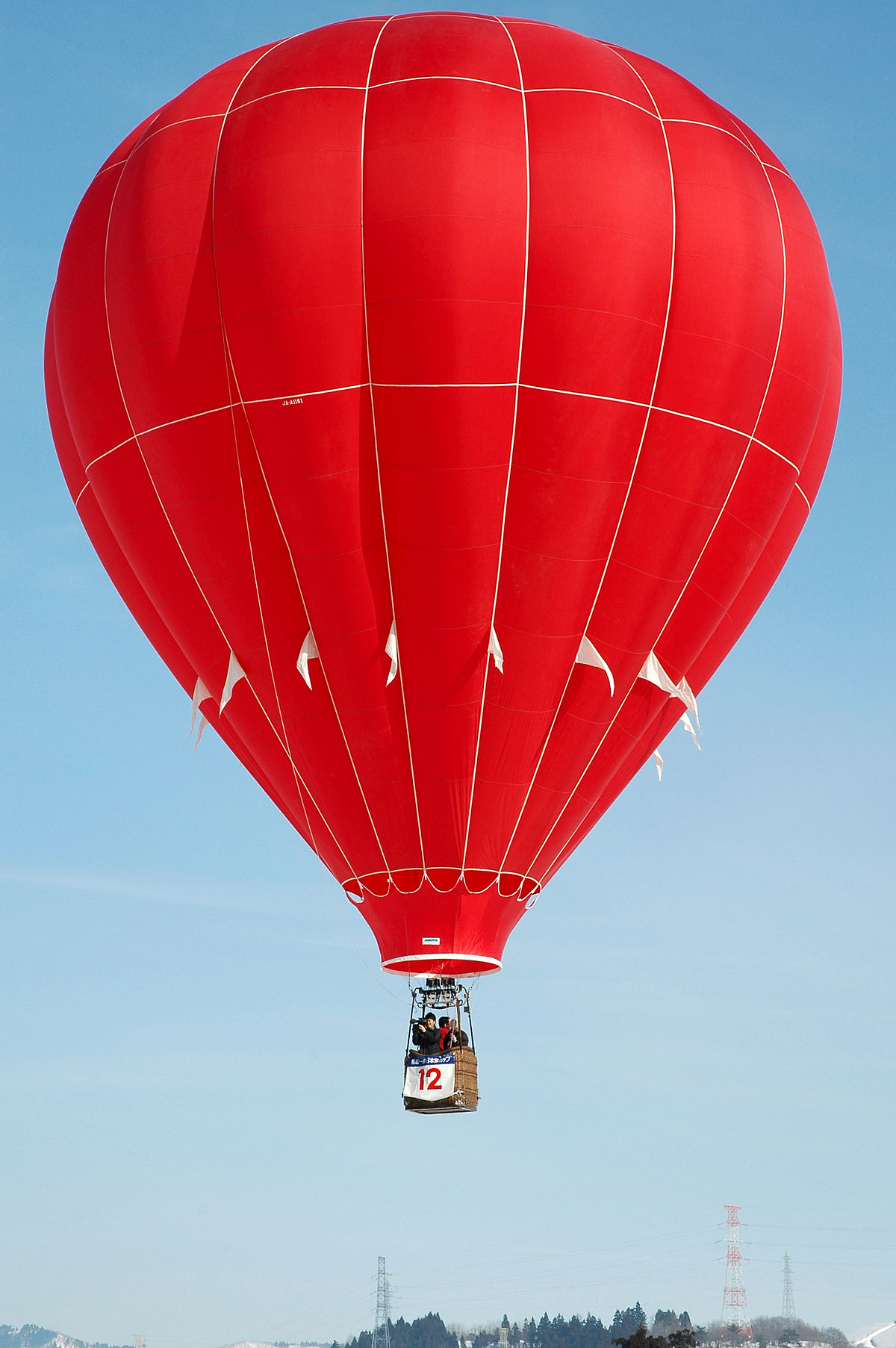}{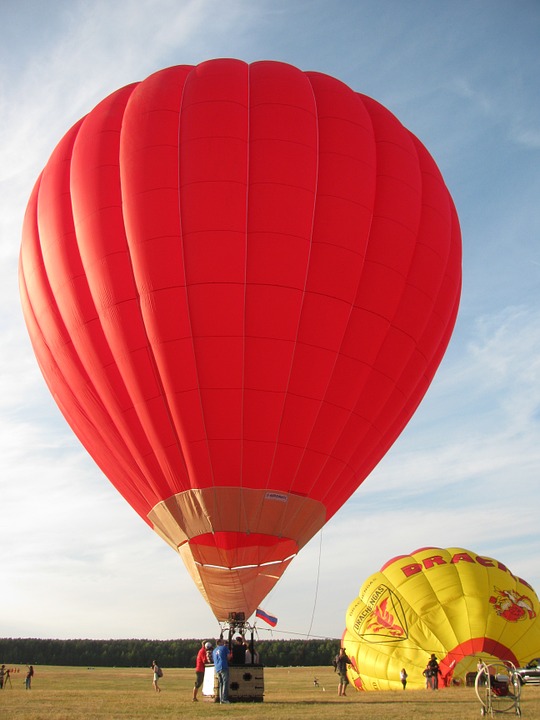} \newline \scriptsize{\textit{Kropsoq (CC BY-SA 3.0); subhv150 (Pixabay)}} & \nlstring{Two hot air balloons are predominantly red and have baskets for passengers.} & $\truelabel$ \\  \dline
     \imgpair{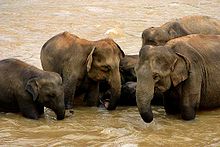}{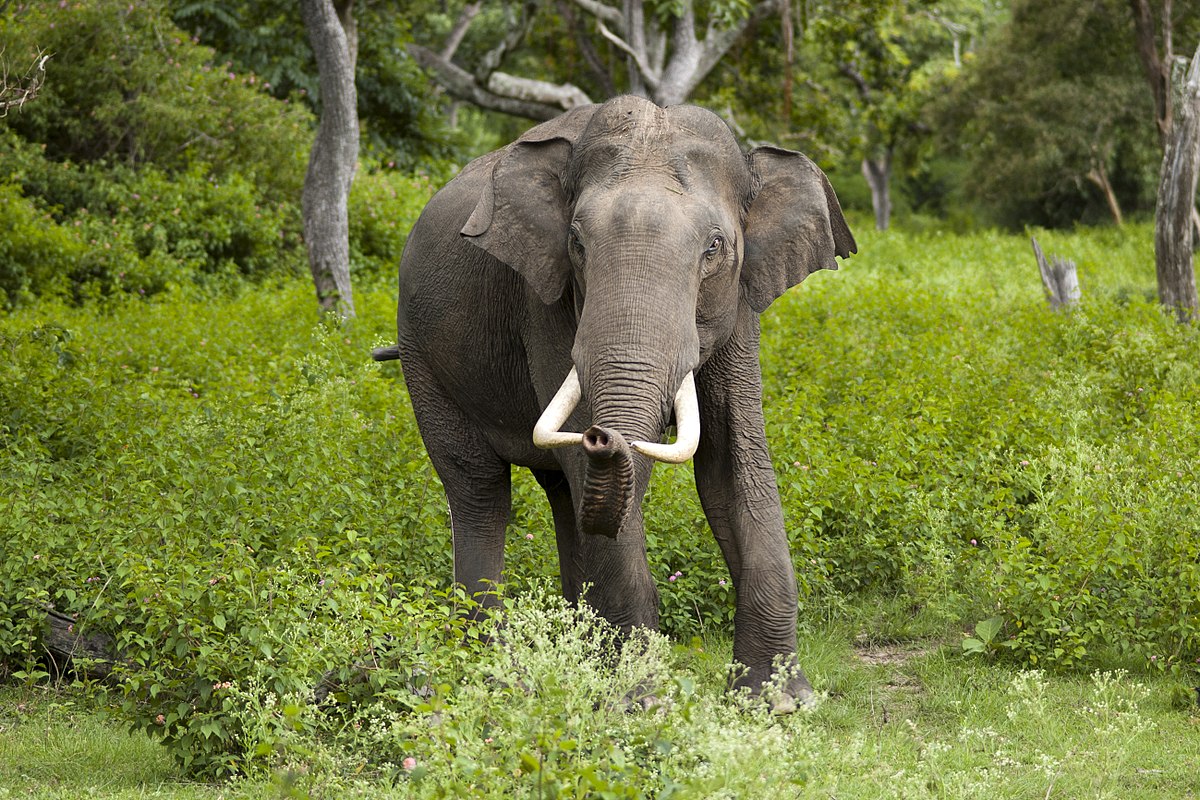} \newline \scriptsize{\textit{babasteve (CC BY 2.0); Yathin S Krishnappa (CC BY-SA 3.0)}} & \nlstring{All elephants have ivory tusks.} & $\falselabel$ \\  \dline
      \imgpair{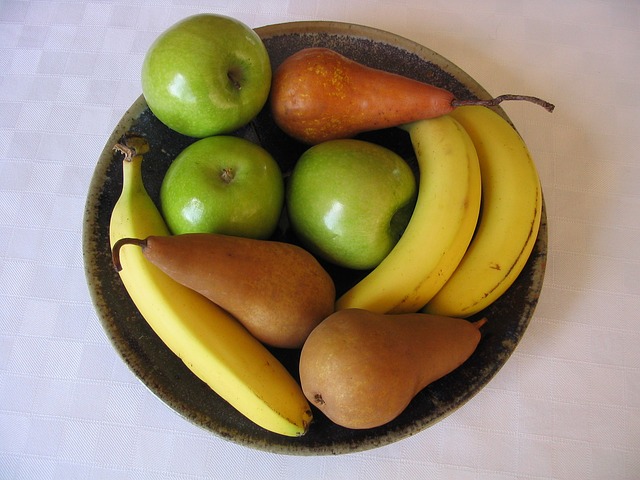}{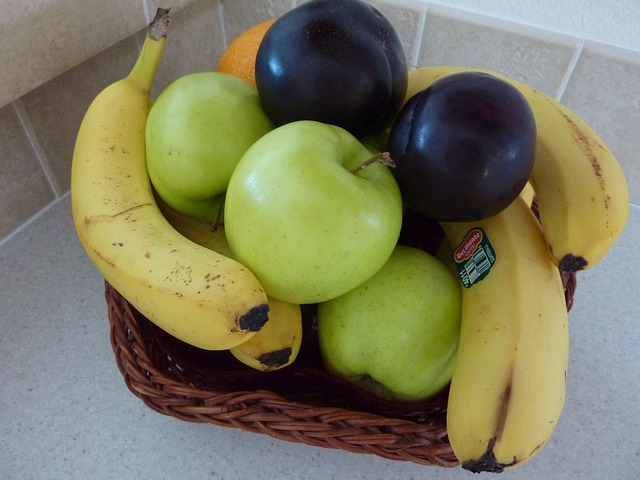} \newline \scriptsize{\textit{NatashaG (Pixabay); Photoman (Pixabay)}} & \nlstring{There are entirely green apples among the fruit in the right image.} & $\truelabel$ \\  \dline
        \imgpair{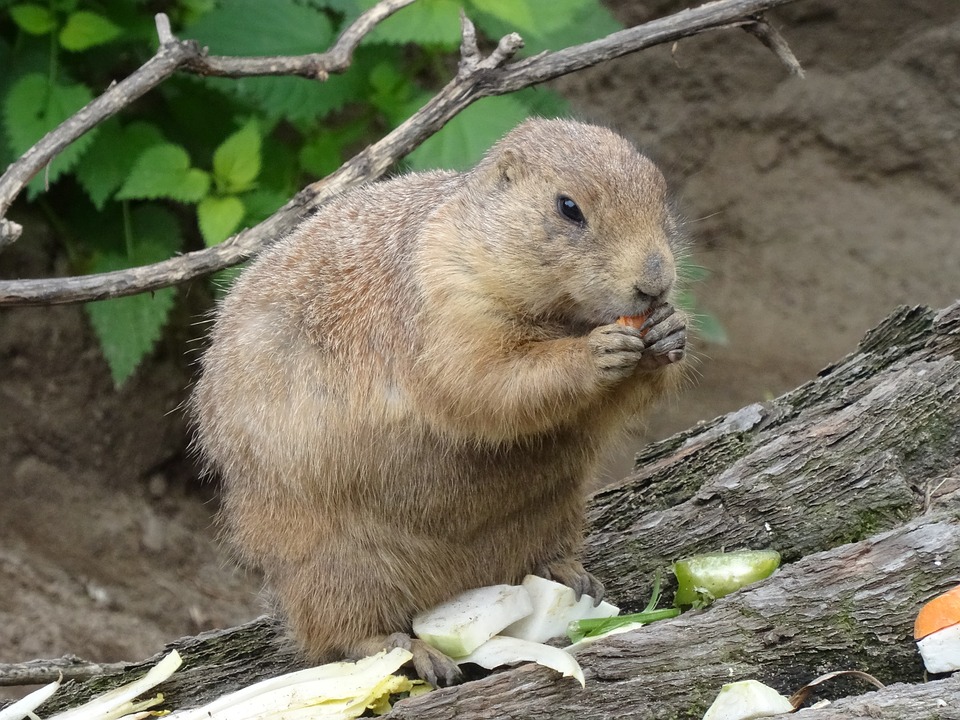}{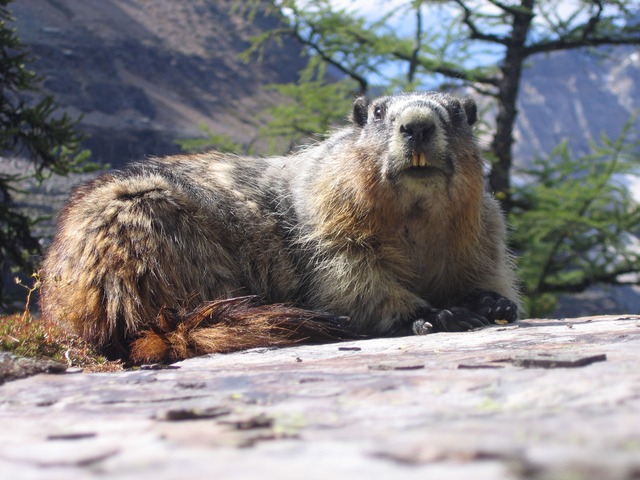} \newline \scriptsize{\textit{Pedi68 (Pixabay); Andrea Schafthuizen (PDP)}} & \nlstring{The animal in the image on the right is standing on its hind legs.} & $\falselabel$ \\  \dline
    \imgpair{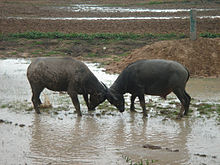}{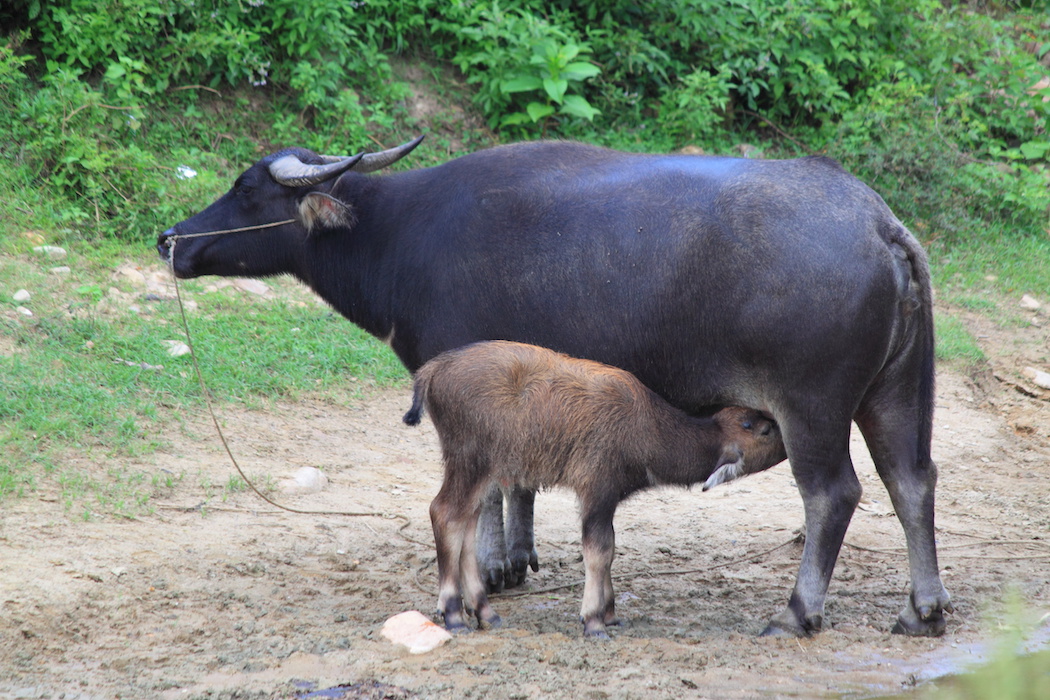} \newline \scriptsize{\textit{Ben \& Katherine Sinclair (CC BY 2.0); Zhangzhugang (CC BY-SA 3.0)}} & \nlstring{One of the images contains one baby water buffalo.} & $\truelabel$ \\  \dline
     \imgpair{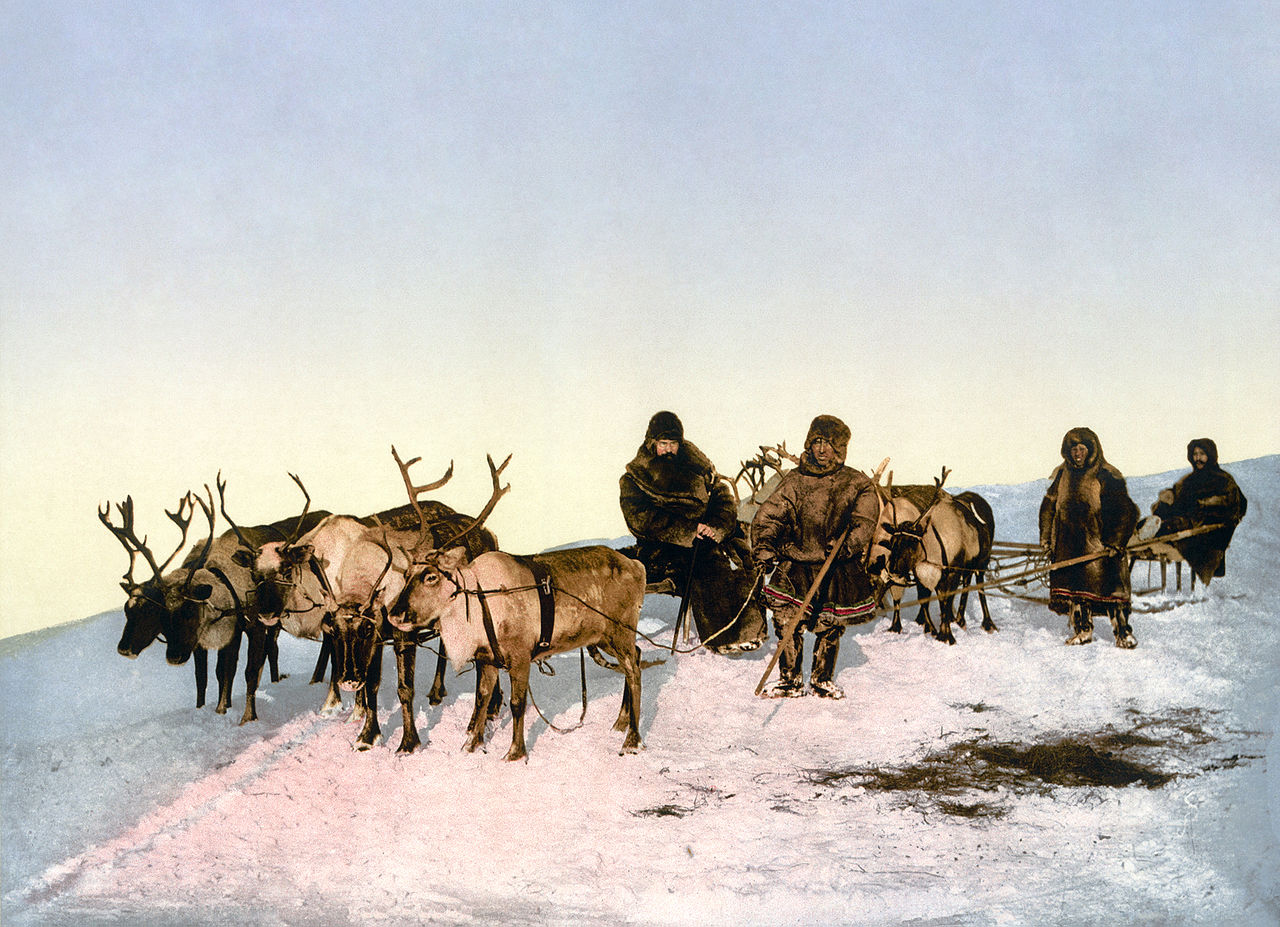}{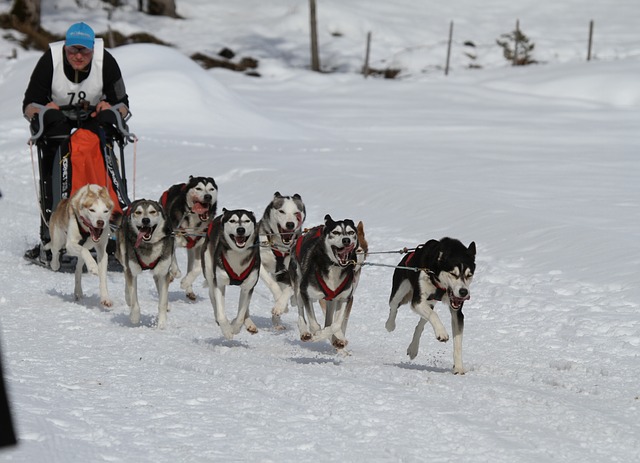} \newline \scriptsize{\textit{Pelikana (CC BY-SA 3.0); violetta (Pixabay)}} & \nlstring{The sled in the image on the left is unoccupied.} & $\falselabel$ \\  \dline
     \imgpair{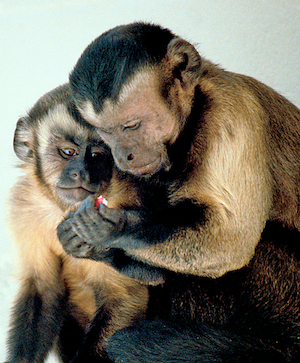}{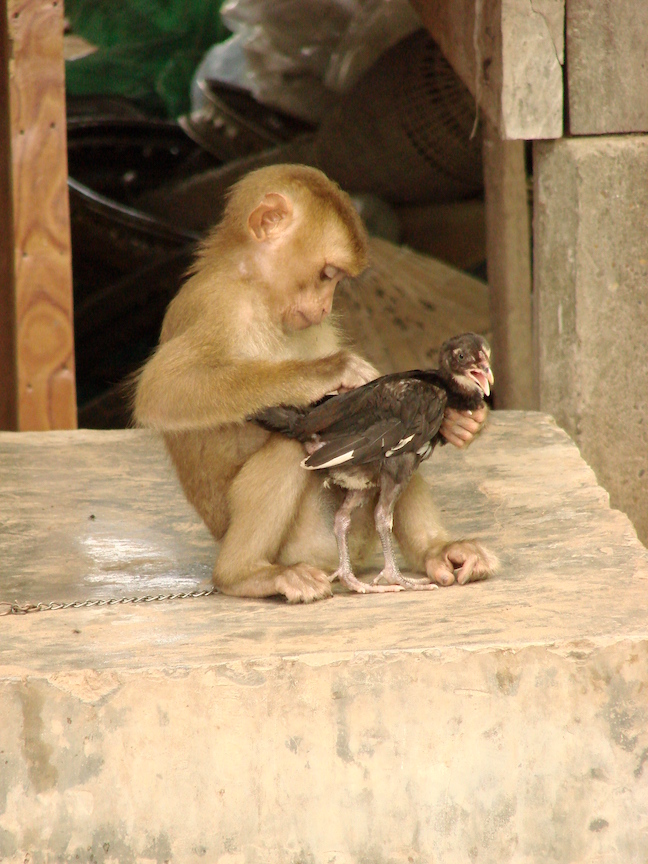} \newline \scriptsize{\textit{Frans de Waal (CC BY 2.5); Adam Jones (CC BY-SA 3.0)}} & \nlstring{Each image shows two animals interacting, and one image shows a monkey grooming the animal next to it.} & $\truelabel$ \\  \dline
     \imgpair{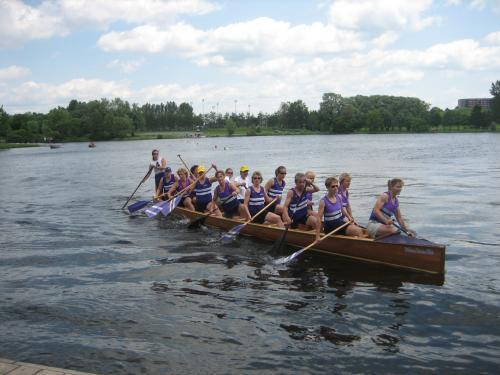}{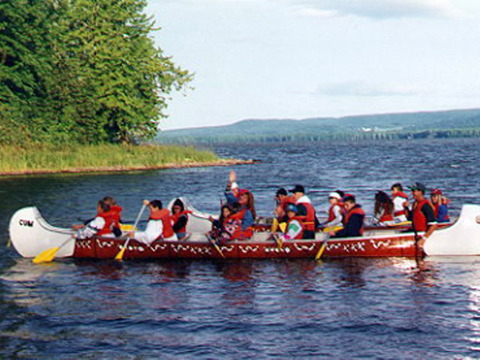} \newline \scriptsize{\textit{Burtonpe (CC BY-SA 3.0); Ville de Montr\'eal (CC BY-SA 3.0)}} & \nlstring{In 1 of the images, the oars are kicking up spray.} & $\falselabel$ \\  \dline
      \imgpair{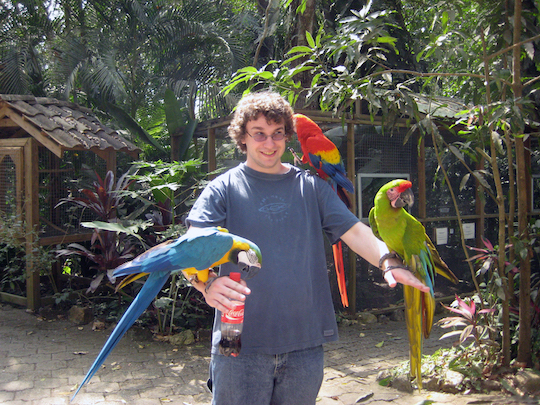}{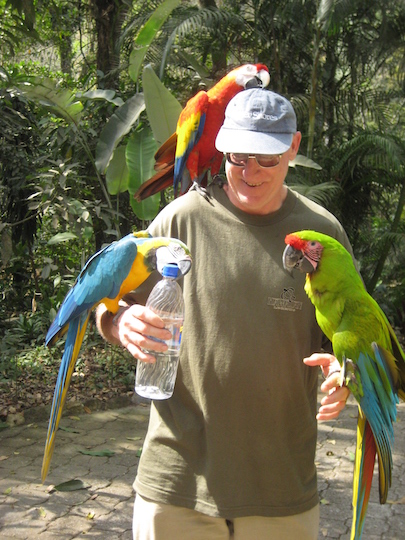} \newline \scriptsize{\textit{Sarah and Jason (CC BY-SA 2.0); Sarah and Jason (CC BY-SA 2.0)}} & \nlstring{In one image, a person is standing in front of a roofed and screened cage area with three different colored parrots perched them.} & $\truelabel$ \\  \dline
  \imgpair{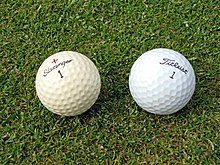}{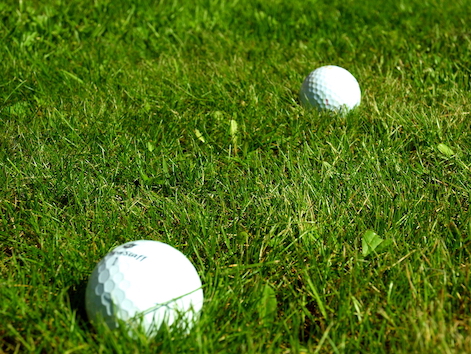} \newline \scriptsize{\textit{Petey21 (CC0); Santeri Viinam\"aki (CC BY-SA 4.0)}} & \nlstring{In one of the images there are at least two golf balls positioned near a hole with a golf flagpole inserted in it.}  & $\falselabel$ \\ 
  \hline
    \end{tabular}
    \caption{Additional examples from the training and development sets of \realnlvr, including license information for each photograph beneath the pair and the label of the example.}
    \label{tab:extra_examples}
\end{table*}